\documentclass[letterpaper, 10 pt, conference]{ieeeconf}  

\IEEEoverridecommandlockouts                              

\usepackage{booktabs}
\usepackage{multirow}

\usepackage{graphicx}
\usepackage{amsmath} 
\usepackage{amssymb}  

\usepackage{caption}

\usepackage[bookmarks=true,colorlinks=true,citecolor=magenta,linkcolor=cyan]{hyperref}
\usepackage{hypcap}

\usepackage{booktabs} 
\usepackage{makecell}
\usepackage{pifont}
\usepackage{float}
\usepackage{subcaption}
\usepackage{array}
\usepackage{siunitx}

\let\labelindent\relax
\usepackage{enumitem}
\usepackage{arydshln}
\usepackage{wrapfig}

\usepackage[nospace]{cite}
\usepackage{ifthen}
\usepackage{xcolor}
\usepackage{multicol}

\newcommand{\xmark}{\ding{55}}
\definecolor{bccolor}{HTML}{E34A6F}
\definecolor{daggercolor}{HTML}{FFBE00}
\definecolor{rlcolor}{HTML}{2398DA}
\definecolor{longcolor}{HTML}{6B529C}
\definecolor{cotrain}{HTML}{39BC56}

\newcommand{\appref}[1]{\hyperref[#1]{App.~\ref{#1}}}
\newcommand{\refeq}[1]{\hyperref[#1]{Eq.~(\ref{#1})}}

\newboolean{draft-mode}
\setboolean{draft-mode}{false}

\newcommand{\anthony}[1]{\ifthenelse{\boolean{draft-mode}}{\textcolor{magenta}{Anthony: #1}}{}}

{}

\newcommand{\lars}[1]{\ifthenelse{\boolean{draft-mode}}{\textcolor{blue}{Lars: #1}}{}}
\newcommand{\idan}[1]{\ifthenelse{\boolean{draft-mode}}{\textcolor{green}{Idan: #1}}{}}
\newcommand{\zw}[1]{\ifthenelse{\boolean{draft-mode}}{\textcolor{purple}{ZW: #1}}{}}
\newcommand{\gmargo}[1]{\ifthenelse{\boolean{draft-mode}}{\textcolor{orange}{Gabe: #1}}{}}

\newcommand{\methodname}{\texttt{ResiP}}
\newcommand{\lamp}{\texttt{lamp}}
\newcommand{\oneleg}{\texttt{one\_leg}}

\newcommand{\roundtable}{\texttt{round\_table}}
\newcommand{\mugrack}{\texttt{mug-rack}}
\newcommand{\peghole}{\texttt{peg-in-hole}}
\newcommand{\bimaninsert}{\texttt{biman-insert}}

\newcommand{\actchunk}{\mathbf{a_t}}

\newcommand{\sethree}{$\text{SE}(3)$}

\newcommand{\xform}{\mathbf{T}}
\newcommand{\argmax}{$\text{argmax}$}

\newcommand{\website}{\href{https://residual-assembly.github.io/}{website}}

\usepackage{balance}

\title{\LARGE \bf
\textit{From Imitation to Refinement} -- Residual RL for Precise Assembly
}

\author{Lars Ankile$^{1,2,3}$ Anthony Simeonov$^{1,2}$ Idan Shenfeld$^{1,2}$ Marcel Torne$^{1,2}$ Pulkit Agrawal$^{1,2}$%
\thanks{$^{1}$Improbable AI Lab $^{2}$Massachusetts Institute of Technology $^{3}$Harvard University {\tt\small ankile@mit.edu}}%
}

\begin{document}

\let\oldtwocolumn\twocolumn
\renewcommand\twocolumn[1][]{%
    \oldtwocolumn[{#1}{
    \vspace{-20pt}
    \begin{flushleft}
           \centering
    \includegraphics[clip,trim=0cm 0cm 0cm 0cm,width=0.99\textwidth]{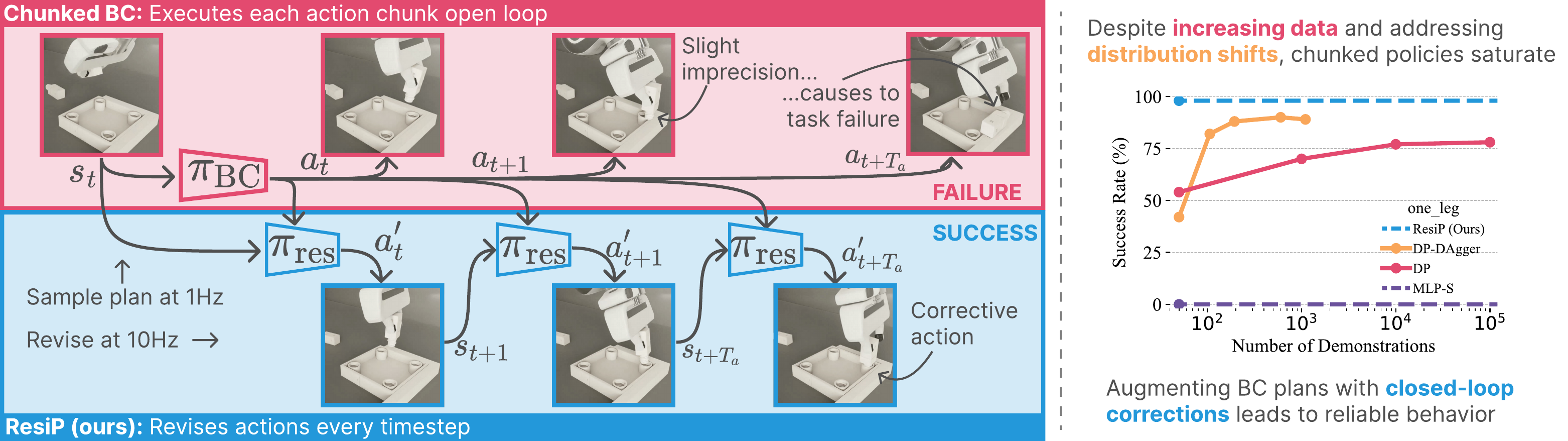}
    \captionof{figure}{%
    \textbf{Left:} \textcolor{bccolor}{(Top)} Tasks like assembly require long-horizon coordination and high-precision control, at which state-of-the-art BC methods fail due to their chunk-level open-loop nature. \textcolor{rlcolor}{(Bottom)} Combining a BC trajectory planner with a closed-loop residual policy trained with RL results in surprisingly robust and reactive behaviors. %
    \textbf{Right:} \textcolor{bccolor}{Chunking} improves performance over \textcolor{longcolor}{standard policy} architectures. Still, performance saturates despite increasing data and addressing distribution shifts with \textcolor{daggercolor}{DAgger}~\cite{ross2011reduction}. Combining chunking with \textcolor{rlcolor}{closed-loop corrections (\methodname{})} combines the strength of each.}\label{fig:overview}
    \end{flushleft}
    }]
}

\maketitle
\thispagestyle{empty}
\pagestyle{empty}

\begin{abstract}
Recent advances in Behavior Cloning (BC) have made it easy to teach robots new tasks. However, we find that the ease of teaching comes at the cost of unreliable performance that saturates with increasing data for tasks requiring precision. The performance saturation can be attributed to two critical factors: (a) distribution shift resulting from the use of offline data and (b) the lack of closed-loop corrective control caused by action chucking (predicting a set of future actions executed open-loop) critical for BC performance.

Our key insight is that by predicting action chunks, BC policies function more like trajectory ``planners'' than closed-loop controllers necessary for reliable execution. To address these challenges, we devise a simple yet effective method, \methodname{} (\textit{Resi}dual for \textit{P}recise Manipulation), that overcomes the reliability problem while retaining BC’s ease of teaching and long-horizon capabilities. \methodname{} augments a frozen, chunked BC model with a fully closed-loop residual policy trained with reinforcement learning (RL) that addresses distribution shifts and introduces closed-loop corrections over open-loop execution of action chunks predicted by the BC trajectory planner.
Videos, code, and data: \href{https://residual-assembly.github.io}{residual-assembly.github.io}. 
\end{abstract}





\section{Introduction}

Many robotic manipulation tasks, such as assembly, require both long-horizon planning and high-precision control, which remains a significant challenge~\cite{kimble2020benchmarking, suarez2016framework, lee2021ikea, heo_furniturebench_2023, zhao_rss23_aloha}. As an example, consider the furniture assembly task depicted in \autoref{fig:overview} (Left), where the robot needs to perform a sequence of steps: grasp, then re-orient the table leg into the correct pose, insert the leg, and finally screw the leg into place. This representative long-horizon task spans several task stages over hundreds of timesteps, each dependent on the successful completion of the previous. During critical moments such as insertions, called ``bottleneck states,'' even slight imprecisions can compound and result in task failure, underscoring the need for reliable execution of each phase.

Behavior cloning (BC) is a popular approach for teaching robots various manipulation skills~\cite{pomerleau1988alvinn, schaal_learning_1996, schaal1999imitation, ratliff2007imitation, agrawal2018computational, zhang2018deep, brohan2022rt, jang2022bc, zhao_rss23_aloha, drolet2024comparison, torne2024reconciling_rialto}. Recent innovations in BC, such as diffusion models~\cite{janner_planning_2022,ajay_is_2023,chi_diffusion_2023,pearce_imitating_2023} and action chunking (predicting a sequence of future actions)~\cite{lai2022action, zhao_rss23_aloha, janner_planning_2022, chi_diffusion_2023}, have enabled learning long-horizon, complex behaviors from demonstrations~\cite{ankile2024juicer, drolet2024comparison}. However, our analysis shows fundamental limitations in BC when applied to tasks requiring high precision: performance saturates with increasing data. For example, the success of a diffusion-based BC model on an insertion task shown in \autoref{fig:overview} (Left) plateaus at $\sim$80\% even with 100,000 demonstrations (\autoref{fig:overview}; right). Recent work finds similar performance saturation~\cite{ross2010efficient, yu2024lucidsim, zhao2024aloha}.

We hypothesize that the performance saturation results from two issues: (i) compounding errors originating from distribution shift as the policy operates on states that increasingly deviate from those seen during training~\cite{ross2010efficient}.
(ii) The use of action-chunking which enables long-horizon control at the cost of reactivity necessary for reliable execution by rolling out each action chunk open-loop~\cite{liu2024bidirectional}. Therefore, we posit that chunked BC policies are best considered ``planners'' rather than reactive controllers. For instance, in the \oneleg{} assembly task shown in \autoref{fig:overview} (left), specific bottleneck states, like insertions, require precise actions at specific time steps. If these critical moments fall within an action chunk, the BC policy cannot make real-time adjustments to compensate for disturbances during execution or inaccuracies in the planned actions.

Reinforcement Learning (RL) is a standard solution to the suboptimal performance of BC polices~\cite{Rajeswaran-RSS-18_dapg, james2022q, lu2022aw_opt, ball2023efficient, schaal1996learning, nair2018overcoming, nakamoto2024cal, uchendu2023jump, hu2023imitation, zheng2022online, kober2010imitation, ramrakhya2023pirlnav, singh2020parrot}. RL fine-tuning effectively overcomes the problem of distribution shifts by generating training data for states visited by the policy. However, recent advancements in BC architectures present new challenges for direct RL fine-tuning. The main issues are that the structure of diffusion models (iterative refinement) and action-chunked policies (resulting in a large action space) make standard RL algorithms unstable (see \autoref{sec:experiments-fine-tuning}) or require significant architectural modifications~\cite{black2023training, drolet2024comparison, ren2024diffusion}.

Instead of invoking RL, one can overcome distribution shift by leveraging an expert and training with supervised learning such as the Dataset Aggregation (DAgger)~\cite{ross2011reduction} algorithm. Our experiments show with roughly ~2000 demonstrations, a DAgger style approach achieves a 90\% success rate on the \oneleg{} task (\autoref{fig:overview})---a significant improvement over pure BC but still 8\% below the expert. One challenge in using DAgger is that it assumes an expert who can be queried on demand, which is usually unavailable. However, another significant challenge is the performance saturation of DAgger, which we attribute to the lack of reactivity because each action chunk is executed in an open-loop fashion. 

\begin{figure}
  \centering
  \includegraphics[width=0.48\textwidth]{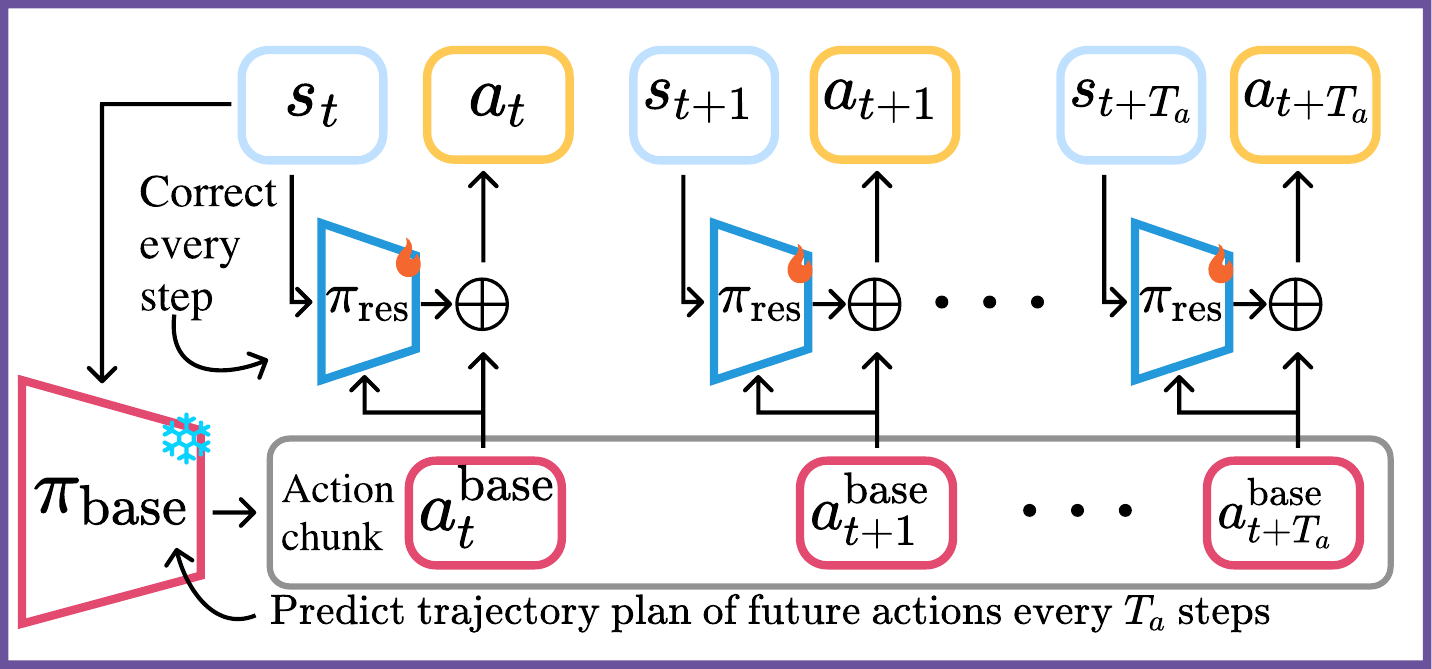}
  \caption{Overview of \methodname{}. A pretrained \textcolor{bccolor}{chunked base policy} predicts an action chunk of $T_a$ future actions. For every timestep, the \textcolor{rlcolor}{residual model} observes the current state $s_t$ and the predicted base action $a_t^{\text{base}}$ and predicts correction.}
  \label{fig:residual-diffusion}
  \vspace{-15pt}
\end{figure}

Our key insight is that action-chunked BC policies function more like ``trajectory planners'' than reactive controllers. To mitigate both distribution shift and lack of closed-loop control, we use a simple method of augmenting a frozen, chunked BC model with a small, single-step residual policy~\cite{silver2018residual, johannink2019residual, alakuijala2021residual, davchev2022residual, carvalho2022residual, shafiullah2022behavior_bet, haldar2023teach_fish_residual, lee2024behavior_vqbet} trained via on-policy RL (\autoref{fig:residual-diffusion}), termed \methodname{} (\textit{Resi}dual for \textit{P}recise Manipulation). Unlike prior residual frameworks, the BC policy generates trajectory segments at coarse temporal resolution ($\sim$1 second), while the residual predicts closed-loop corrections at each control step ($\sim$0.1 second). 

\methodname{} addresses both challenges faced by BC methods: the RL-trained residual overcomes distribution shifts and by closing the sensorimotor loop more frequently it lends reactivity by enabling precise adjustments that chunked policies cannot provide. Using only a sparse task-completion reward, this approach achieves a 98\% success rate starting with just 50 expert demonstrations on the \oneleg{} task---an 18 percentage point improvement over BC with 100K demonstrations.

Since our method relies on training RL policies, our learning pipeline heavily relies on simulation. Applying our pipeline requires constructing a digital twin of the real-world use case (i.e., real-to-sim~\cite{torne2024reconciling_rialto}), training with \methodname{}, and finally transferring the trained policy to the real world. While our focus in this paper is on \methodname{} and highlighting the performance saturation of purely BC methods, our experiments demonstrate the entire pipeline on several challenging assembly tasks.

\section{Method}

Our method applies to any task that can be simulated using already available assets (e.g., CAD models) or by scanning real-world objects~\cite{torne2024reconciling_rialto, villasevil2024scaling}. While this paper focuses on assembly tasks for which CAD models are already available for constructing simulation environments, our method applies to many tasks of practical interest. Note that our method requires CAD models only for training and not during deployment.

\begin{figure*}[t]
    \centering
    \includegraphics[width=\linewidth]{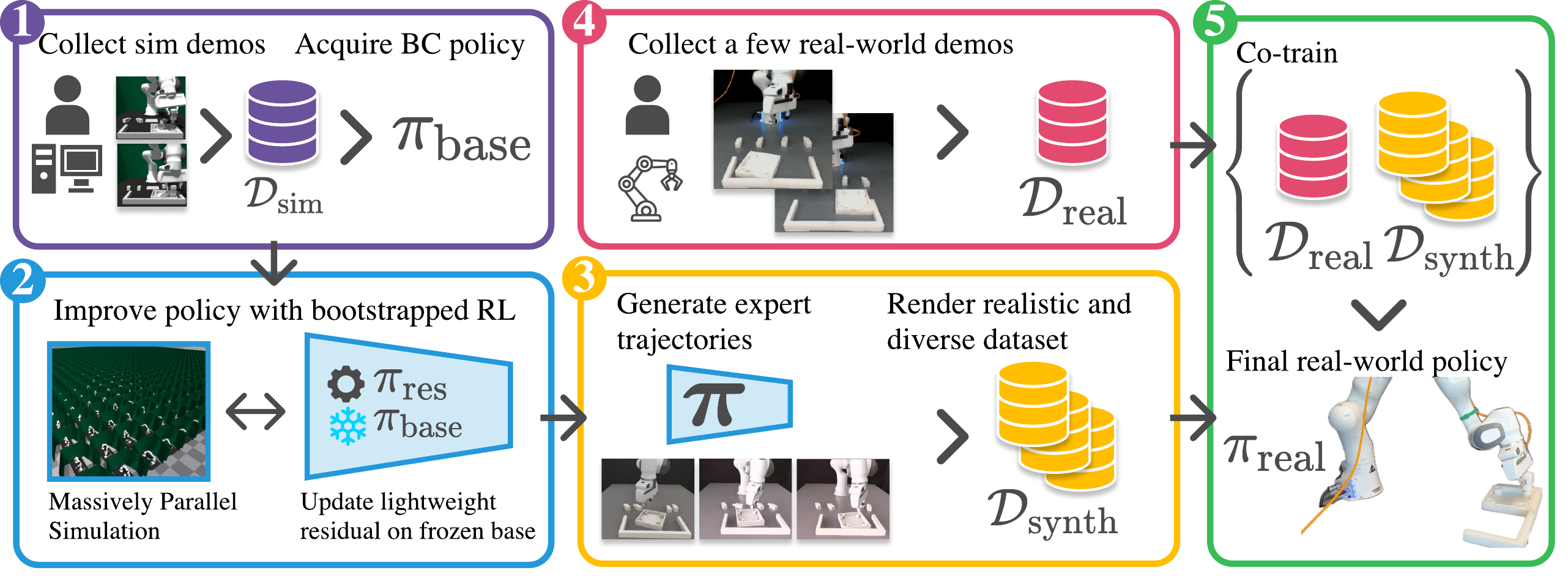}
    \caption{\textbf{Sim-to-real pipeline.} \textcolor{longcolor}{\textbf{(1)}} Beginning with a policy trained with BC in simulation, \textcolor{rlcolor}{\textbf{(2)}} we train residual policies with RL and sparse rewards. \textcolor{daggercolor}{\textbf{(3)}} We then distill the resulting behaviors into a policy operating from RGB images. \textcolor{bccolor}{\textbf{(4)}} By combining synthetic data with a small set of real demonstrations, \textcolor{cotrain}{\textbf{(5)}} we deploy RGB-based policies in the real world.}
    \label{fig:pipeline}
    \vspace{-15pt}
\end{figure*}

Our approach from task definition to deployable vision policy consists of three key components (see \autoref{fig:pipeline} for an overview). First, we train a base policy using behavior cloning (BC) on a small set of demonstrations (\autoref{sec:behavior-cloning}) in simulation. Then, we improve this policy's precision by training a residual component with reinforcement learning that makes closed-loop corrections to the base policy's actions (\autoref{sec:reactive-control}). Finally, we learn a real-world deployable policy with policy distillation and co-training with a few real-world demonstrations (\autoref{sec:sim-to-real-pipeline-method}).

\subsection{Problem Formulation}
\label{sec:problem-formulation}

We formulate the target task as a discrete-time sequential decision-making problem. At each timestep $t$, the robot receives an observation $o_t \in \mathcal{O}$ corresponding to the underlying state $s_t \in \mathcal{S}$ of the robot and environment. Given the observation, the policy outputs an action $a_t \in \mathcal{A}$ that is executed in the environment. The action space $\mathcal{A}$ consists of end-effector poses in $\sethree{}$ and gripper commands. The state space includes the robot's configuration (end-effector pose, velocity, and gripper state) and the poses of all objects in the scene. Task completion is defined through sparse task completion rewards. We define this reward as an instantaneous signal that provides 1 at the timestep when a pair of objects achieves a pre-defined relative pose corresponding to successful assembly and 0 otherwise, including after the alignment is maintained. See more details about the environments in \appref{app:tasks-details}.

\subsection{Base Policy Learning via Behavior Cloning}
\label{sec:behavior-cloning}

For each task, we collect a dataset of 50 demonstrations in simulation by teleoperating the robot, $\mathcal{D}_{\text{sim}}:\ \{\tau_1, ..., \tau_N\}$. Each trajectory, $\tau$, contains system states $s_{t}$, and robot actions $a_{t}$, i.e., $\tau_i=\{(s_t, a_1), ..., (s_T, a_T)\}$, with $T$ being the trajectory length. We use $\mathcal{D}_{\text{sim}}$ to first train base policy $\pi_{\text{base}}$ with Behavior Cloning (BC), i.e., $\pi_{base} =\ \argmax_{\pi_{\text{base}}}\ \mathbb{E}_{(a_t,s_t)\sim\mathcal{D}_{\text{sim}}} \left[\log\pi_{\text{base}}(a_t|s_t) \right]$ using the Diffusion Policy (DP) architecture~\cite{chi_diffusion_2023}, which serves as the starting point for Reinforcement Learning (RL). Consistent with state-of-the-art in BC training~\cite{chi_diffusion_2023, zhao_rss23_aloha}, we use action chunks that predict a set of multiple future actions instead of a single action at every timestep. This chunking approach effectively reduces the task horizon, enabling better performance on long-horizon tasks, but as we show (\autoref{sec:resip-ablations}), it sacrifices the ability to make closed-loop corrections. We denote the length of future action sequences predicted by the policy as $T_a$, the output as $\actchunk = [a_t^{\text{base}}, ..., a_{t+T_a}^{\text{base}}]$. When predicting an action chunk $\actchunk$ of length $T_a$, we only execute a subset $[a_{t}^{\text{base}},...,a_{t+T_{\text{exec}}}^{\text{base}}]$, with execution horizon $T_{\text{exec}} \leq T_a$. Similar to~\cite{chi_diffusion_2023}, we use $T_{\text{exec}} = 8$, and $T_a = 32$ instead of their 16 as we empirically found it performs better.

\subsection{Reactive Control via \methodname{}}
\label{sec:reactive-control}

Given the initial chunked base policy $\pi_{\text{base}}$ obtained by BC described above, we want to improve the policy to overcome the issues of distribution shift and the lack of reactivity.
One way to mitigate the adverse effects of distribution shifts is to fine-tune the BC-trained policy with RL~\cite{pomerleau1988alvinn, schaal_learning_1996, schaal1999imitation, ratliff2007imitation, agrawal2018computational, zhang2018deep, brohan2022rt, jang2022bc, zhao_rss23_aloha, drolet2024comparison, torne2024reconciling_rialto}. We side-step the complications of direct RL fine-tuning of action-chunked policies (see discussion in \autoref{app:tasks-details}) by training a residual~\cite{silver2018residual, johannink2019residual, alakuijala2021residual, davchev2022residual, carvalho2022residual, shafiullah2022behavior_bet, haldar2023teach_fish_residual, lee2024behavior_vqbet} Gaussian Multi-Layer Perceptron (MLP)~\cite{rumelhart1986learning} policy $\pi_{\text{res}}$ using PPO~\cite{schulman_proximal_2017}. Our key design choice is that while this residual framework can address distribution shift through either chunk-level or timestep-level corrections (see \autoref{sec:closed-loop-ablation} for analysis), the latter enables reactivity helpful for precise manipulation.

For each timestep $i = 0, ..., T_a - 1$ within the base policy's action chunk, we form the residual's observation by concatenating the full simulation state (robot proprioceptive information and object poses) with the base policy's predicted action, $s_{t+i}^\text{res}=[s_{t+i},a^\text{base}{t+i}]$. The residual policy then produces a corrective action $a_{t+i}^{\text{res}} \sim \pi_{\text{res}}(\cdot | s_{t+i}^\text{res})$ that modifies the base action: $a_{t+i} = a_{t+i}^{\text{base}} + a_{t+i}^{\text{res}}$. The resulting fine-tuned policy is a combination of the pre-trained BC policy $\pi_\text{base}$ and the correction policy $\pi_\text{res}$, denoted $\boldsymbol{\pi}$. As demonstrated in \autoref{sec:closed-loop-results}, this per-timestep correction capability is crucial for achieving reliable performance in precision-critical tasks.

During training, as in any RL fine-tuning, one needs to adjust the exploration noise to provide sufficient stochasticity to discover new behaviors while maintaining enough precision for task success (see \appref{app:alpha-ablation} for detailed analyses of design choices). The residual policy uses orthogonal initialization~\cite{saxe2013exact} with a small gain factor on the final layer, ensuring uncorrelated initial corrections centered around zero---a design choice that aligns with our assumption that the base policy's errors are unbiased in expectation. Since most task failures stem from slight imprecisions in the base policy's actions, this architecture naturally leads to learning small, targeted corrections around the base policy's behavior.

\begin{figure}
    \vspace{5pt}
    \centering
    \includegraphics[width=0.98\linewidth]{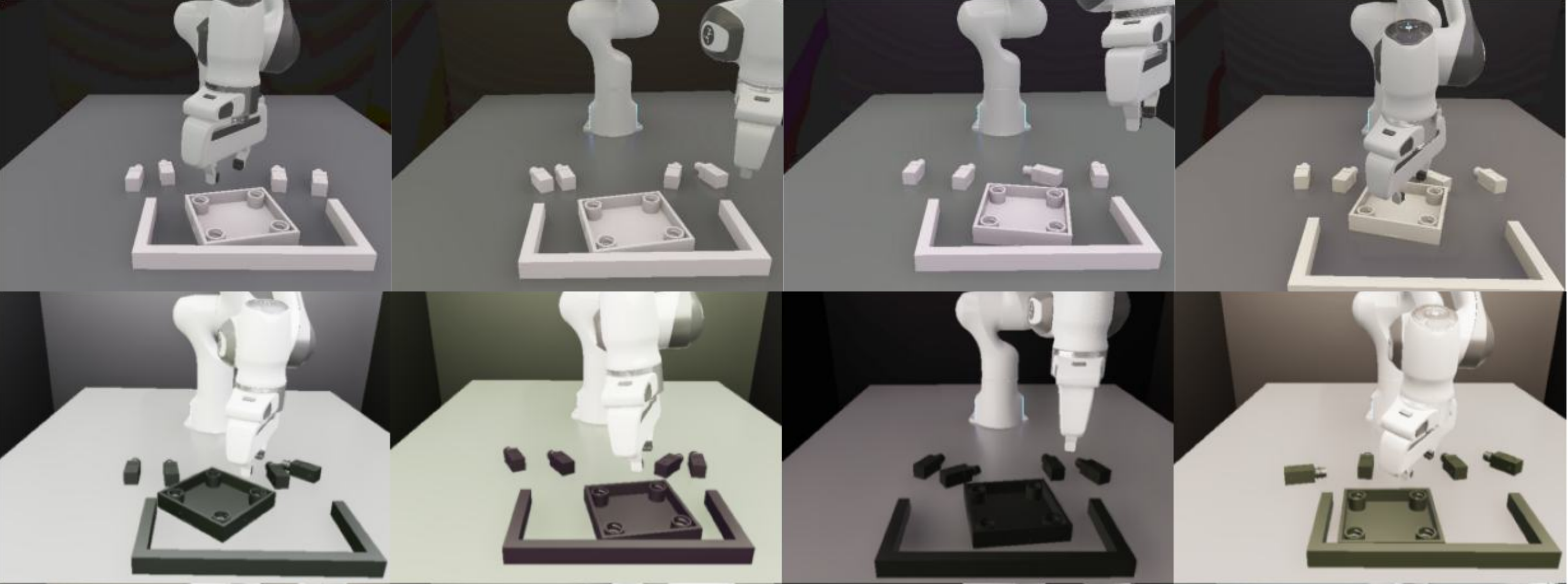}
    \caption{Photorealistic rendering and domain randomization for sim-to-real transfer. In the first row, we show our nominal experiment setting of rendering parts in the original white color while varying the position, brightness, and hue of the lighting in the scene. In the bottom row, we also introduce variations in the part colors to see how that can help the final policy adapt to unseen part appearances without collecting any additional data.}
    \label{fig:synthetic-data}
    \vspace{-15pt}
\end{figure}

The policy observation for simulation training, $\mathcal{S}$ contains the 6 DoF end-effector pose $\xform$, spatial velocity $\mathbf{V}$, and gripper width $w_{\text{g}}$, along with the 6-DoF poses of all the parts in the environment $\{\xform{}^{\text{part}_{i}}\}_{i=1}^{\text{num\_parts}}$.

\begin{figure*}[t]
    \centering
    \includegraphics[width=0.98\linewidth]{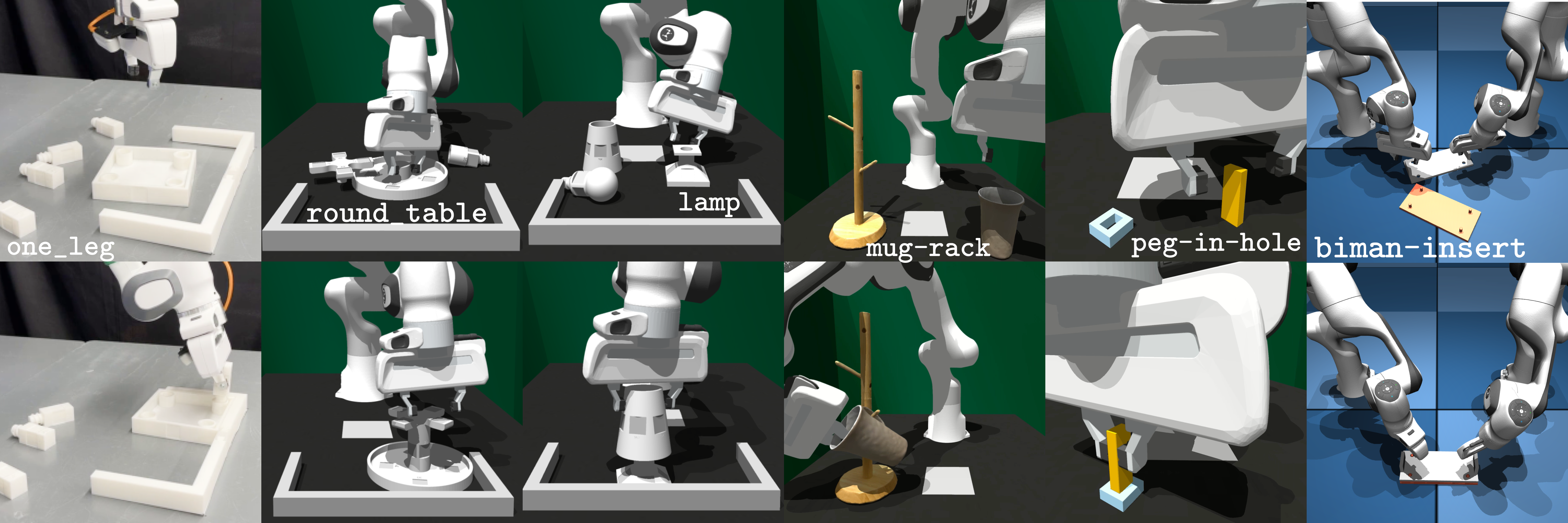}
    \caption{Initial and goal states for our 6 tasks. The first three tasks, \oneleg{}, \roundtable{}, and \lamp{}, are from the FurnitureBench~\cite{heo_furniturebench_2023} task suite. \mugrack{} we created by scanning objects and importing them into the simulation. \peghole{} is from the Factory~\cite{yash_factory2022_rss} task suite. \bimaninsert{} we created using the simulation-based teleoperation system DART~\cite{park2024dexhub}. Our task suite exhibits diverse challenges like long horizons, tight tolerances, bimanual control, and multi-modal success criteria.}
    \label{fig:task-overview}
    \vspace{-15pt}
\end{figure*}

\subsection{Sim-to-Real Transfer}
\label{sec:sim-to-real-pipeline-method}

We treat $\pi_{res}$ trained in simulation using privileged state information, $\mathcal{S}$, as a teacher policy $\pi_\text{teacher}$ to distill into a student policy $\pi_\text{student}$ that takes as input sensory observations and can therefore be deployed in the real world~\cite{lee2020learning, kumar2021rma, chen2023visual, torne2024reconciling_rialto}. The real world observation space $\mathcal{O}$ contains the robot end-effector pose $\xform{} \in $ \sethree{}, robot end-effector spatial velocity $\mathbf{V} \in \mathbb{R}^{6}$, the gripper width $w_{\text{g}}$, and RGB images from a fixed front-view camera ($I^{\text{front}} \in \mathbb{R}^{h \times w \times 3}$) and a wrist-mounted camera ($I^{\text{wrist}} \in \mathbb{R}^{h \times w \times 3}$), each with uncalibrated camera poses.

For teacher-student distillation, we collect a dataset of successful trajectories $\mathcal{D}_{\text{synth}}=\{\tau_{\text{synth},1}, ..., \tau_{\text{synth},N}\}$, $\tau_{\text{synth},i}=\{(s_1, a_1), ..., (s_T, a_T)\}$, from $\pi_{teacher}$ initialized with a diverse set of initial object poses.
To bridge the sim-to-real gap, we convert the trajectory dataset with only state information ($\mathcal{S}$) into trajectories with sensory observations ($\mathcal{O}$) by re-rendering the trajectories into realistic-looking image observations (see \autoref{fig:synthetic-data} for examples). This rendering process allows us to introduce visual variations---including object colors, textures, lighting conditions, and camera perspectives---that are not easily obtainable with standard image augmentations or real-world data collection (see \appref{app:sim-to-real-analysis} for examples and details). We denote this rendered dataset as $\mathcal{D}_{\text{synth-render}}$.

$\pi_\text{student}$ is represented with a Diffusion Policy architecture that uses ResNet18~\cite{he2016deep} vision encoder pre-trained on robotic manipulation data~\cite{nair_r3m_2022} (see \appref{app:rgb-policy-implementation-details} for implementation details). To minimize the sim-to-real gap, we combine a small set of real-world demonstrations $\mathcal{D}_{\text{real}}$ (10-40 trajectories) collected directly on the robot with synthetic rendered trajectories from $\mathcal{D}_{\text{synth-render}}$ (approximately 400 demonstrations)~\cite{torne2024reconciling_rialto}. This synthetic-to-real data ratio provides increased domain coverage while not washing out real-world demonstrations. The real demonstrations contain only RGB observations and no ground-truth pose information. This combined dataset $\mathcal{D}_{\text{real}} \cup \mathcal{D}_{\text{synth-render}}$ is used to train the student policy $\pi_\text{student}$ with the same BC loss as in the simulation-based experiments.


\section{Experimental Setup}

\subsection{Tasks and Environment}

We evaluate our method on a set of challenging assembly tasks of \oneleg{}, \roundtable{}, and \lamp{} from the FurnitureBench~\cite{heo_furniturebench_2023} task suite, the \peghole{} task from~\cite{yash_factory2022_rss}, a custom mug-hanging task we call \mugrack{}, and a custom bimanual precise insertion task we call \bimaninsert{}. Examples of all tasks are shown in \autoref{fig:task-overview}. Visualizations of initial state distributions and detailed task descriptions are provided in \appref{app:tasks-and-environment}, and task rollouts can be seen on the \website{}.
 
Multi-part tight-tolerance assembly interactions are simulated using the SDF-based collision geometry representations provided as part of the Factory~\cite{yash_factory2022_rss} extension of Nvidia's IsaacGym simulator~\cite{makoviychuk2021isaac}, with demos collected using a \href{https://3dconnexion.com/us/product/spacemouse-wireless/}{SpaceMouse}. The \bimaninsert{} task is implemented using the MuJoCo simulator~\cite{todorov2012mujoco}, with the demos provided using the DART teleoperation system from~\cite{park2024dexhub}.

We define a simple, sparse task-completion reward set to 1 for each task when a pair of parts has been fully assembled and 0 otherwise. For example, in the \lamp{} task, the policy receives two binary rewards: the first when the bulb is fully screwed in and the second when the lamp shade is successfully placed. Some of our tasks are long horizon (up to $\sim$750-1000 steps at 10Hz) and require sequencing of behaviors such as 6-Degree-of-Freedom (DoF) grasping, reorientation, insertion, and screwing (see \autoref{fig:overview}; Left).

\begin{table*}
    \centering
    \resizebox{\textwidth}{!}{
    \begin{tabular}{@{}c@{}lccccccccc@{}}
        \toprule
        & \multirow{2}{*}{Methods} & \multicolumn{2}{c}{\oneleg{}} & \multicolumn{2}{c}{\roundtable{}} & \multicolumn{2}{c}{\lamp{}} & \multicolumn{1}{c}{\mugrack{}} & \multicolumn{1}{c}{\peghole{}} & \multicolumn{1}{c}{\bimaninsert{}} \\
        \cmidrule(lr){3-4} \cmidrule(lr){5-6} \cmidrule(lr){7-8} \cmidrule(lr){9-9} \cmidrule(lr){10-10} \cmidrule(lr){11-11}
        & & Low & Med & Low & Med & Low & Med & Low & Low & Low  \\
        \midrule
        \multirow{3}{*}{\rotatebox[origin=c]{90}{\textbf{BC}}\hspace{0.5em}}
        & MLP-S & 0  &  0 & 0 & 0 & 0 & 0 & 0 & 2 & 0 \\
        & MLP-C & 45 & 10 & 5 & 2 & 8 & 1 & 21 & 2 & 7 \\
        & DP   & \textbf{54} & \textbf{26} & \textbf{12} & 4 & 7 & 2 & \textbf{26} & \textbf{5} & \textbf{33} \\
        \midrule
        \multirow{3}{*}{\rotatebox[origin=c]{90}{\textbf{RL}}\hspace{0.5em}}
        & PPO-C & 70 & 28 & 38 & 6 & 32 & 2 & 23 & 4 & 30 \\
        & IDQL & 57 & 27 & 18 & 3 & 11 & 1 & 31 & 3 & 40 \\
        & \methodname{} (ours) & \textbf{98} & \textbf{76} & \textbf{94} & \textbf{77} & \textbf{97} & \textbf{70} & \textbf{88} & \textbf{99} & \textbf{93} \\
        \bottomrule
    \end{tabular}
    }
    \caption{Success rates (percentage of successful completions over 1024 evaluation episodes using each method's best-performing checkpoint) for different policy architectures across our task suite. \textbf{Top:} Baseline BC methods, where Diffusion Policies (DP) generally outperform MLPs with chunking (MLP-C), while MLPs without chunking (MLP-S) are unable to solve all tasks except \peghole{}. \textbf{Bottom:} RL-based methods, where our proposed residual policy (\methodname{}) provides significant improvements in success rates over the base BC policy, improvements not seen from the alternative RL methods.}
    \label{tab:performance_comparison}
    \vspace{-15pt}
\end{table*}

\subsection{System Configuration}

The policy operates at 10Hz on a 7 Degrees-of-Freedom (DoF) Franka Emika Panda robot arm for all tasks but \bimaninsert{}, which operates at 50Hz on two Franka Panda arms (i.e., 14 DoF). The action space consists of the desired end-effector pose $\xform{}^{\text{des}} \in$ \sethree{} (i.e., both position and orientation) and a binary gripper command for opening/closing the parallel-jaw gripper. These desired end-effector poses are converted to joint position targets using differential inverse kinematics~\cite{manipulation}, then tracked using a low-level PD controller running at 1KHz with manually specified stiffness and dampening parameters lightly tuned to balance compliance with accurate trajectory tracking.

\subsection{Evaluation protocol}
\label{sec:evaluation-protocol}

\subsubsection{Primary Evaluation}

We evaluate all methods by executing the policy from an initial state sampled from the same initial state distribution used for collecting demonstrations. We follow the default randomization protocol used in FurnitureBench~\cite{heo_furniturebench_2023}: at the beginning of each episode, objects are randomly offset from their nominal positions by $\Delta x, \Delta y \in [-1.5, 1.5]$ cm in the horizontal plane (with fixed height $z$) for low randomization settings and by $\Delta x, \Delta y \in [-5, 5]$ cm for medium randomization. For the U-shaped fixture used to stabilize parts during assembly (the three-sided white ``wall'' surrounding the parts in the left-most three columns in \autoref{fig:task-overview}), we apply additional offsets of $\Delta x, \Delta y \in [-2, 2]$ cm and $\Delta x, \Delta y \in [-4, 4]$ cm from its nominal position for low and medium settings respectively. 

We define task success as achieving the target geometric alignment between pairs of parts, as specified by the task reward function in \autoref{sec:problem-formulation}. For each method, we report the success rate calculated over 1024 evaluation episodes using the best-performing checkpoint, as our goal is to demonstrate the potential capabilities of each architectural approach rather than average performance across training runs.

\subsubsection{Robustness to Dynamic Disturbances}

To further evaluate robustness to dynamic disturbances, we also perform a version of the above evaluation that includes random force perturbations during policy execution. At each timestep, we randomly sample 1\% of the objects in the environment and apply a perturbation corresponding to the same randomization described above for initialization. These perturbations simulate unexpected contact forces and dynamic disturbances not seen during training. We evaluate each method's performance degradation under these conditions by comparing success rates with/without perturbations across 1024 rollouts.

\subsubsection{Real-World Evaluation Protocol}

For sim-to-real transfer, we use IsaacSim~\cite{mittal2023orbit} to render photorealistic trajectories of our simulation data, as it provides better rendering capabilities than IsaacGym~\cite{makoviychuk2021isaac}. We evaluate policies across a grid of initial object and obstacle poses for real-world experiments that match our low randomization simulation setting ($\Delta x, \Delta y \in [-1.5, 1.5]$ cm). We perform 10 trials for each method and ensure that each method is tested on the same initial object states. Success criteria remain consistent with our simulation experiments: achieving target geometric alignment between assembly parts.

\subsection{Baselines and Ablations}
\label{sec:baselines}

We evaluate our method against several baselines to analyze the importance of action chunking, policy architecture, and closed-loop control learned with Reinforcement Learning (RL). Implementation details and hyperparameters are provided in \appref{sec:implementation-details}.

\subsubsection{Behavior Cloning Baselines}

We first evaluate the impact of action chunking by comparing a standard Behavior Cloning (BC) approach using an MLP (MLP-S) with a version that predicts action chunks (MLP-C). We find the Diffusion Policy architecture~\cite{chi_diffusion_2023} provides the strongest BC performance and use it as both our primary baseline and the foundation for \methodname{}.

\subsubsection{Distribution Shift Analysis}

To assess the impact of distribution shifts, we implement \textbf{DP-DAgger}~\cite{ross2011reduction}, which iteratively queries an expert policy to gather corrective demonstrations in states visited by the learned policy. Our experiments use \methodname{} as the expert. The DAgger policy is trained with the same BC loss and architecture as the nominal DP policy.

\subsubsection{Reinforcement Learning Comparisons}

Building on our BC policies, we compare two approaches for RL fine-tuning. First, we evaluate PPO fine-tuning of our chunked MLP policy (PPO-C)~\cite{schulman_proximal_2017}, treating each action chunk as a single concatenated action. For our diffusion-based policy, we implement IDQL~\cite{hansen-estruch_idql_2023}, where multiple action chunks are sampled from the diffusion policy and selected based on learned Q-values using the on-policy method of~\cite{shenfeld2024value}.

\begin{figure}
    \centering
    \includegraphics[width=0.98\linewidth]{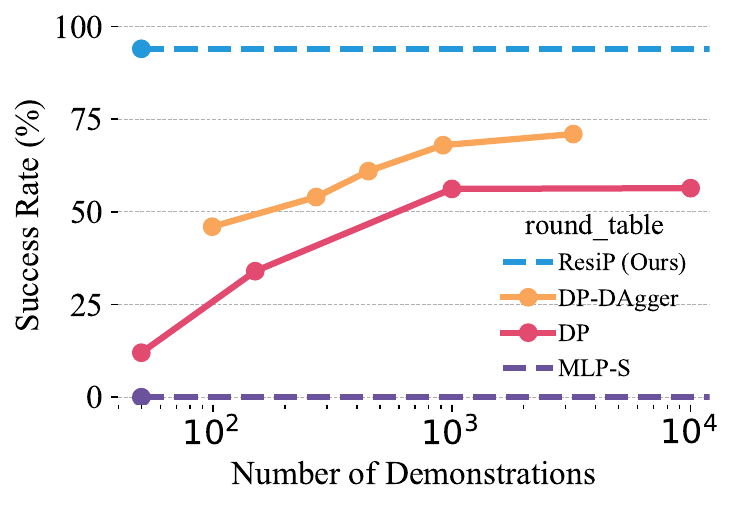}
    \caption{Scaling behavior comparison on the harder \roundtable{} task mirrors the trends observed for \oneleg{} (\autoref{fig:overview} (right)), but with lower overall performance. Just as in \oneleg{}, increasing the number of demonstrations for BC training leads to diminishing returns, saturating at 56\% success rate. While DAgger's online data collection performs better, it also plateaus at 71\%, well below \methodname{}'s 94\% success rate. This consistent pattern across tasks, with lower saturation points for the more complex tasks, highlights the limitations of offline approaches and the benefits of learning closed-loop control online.}
    \label{fig:scaling-round-table}
    \vspace{-15pt}
\end{figure}

\subsubsection{Closed-Loop Control Ablation}
\label{sec:closed-loop-ablation}

To study the importance of per-timestep corrections, we implement \textbf{\methodname{}-C}, a variant of our method that predicts residual corrections at the same frequency as the base policy's action chunks. While the standard \methodname{} observes the current state and predicts a correction for each timestep, \methodname{}-C observes the current state and all actions in the predicted chunk and predicts a correction of all actions in the chunk. \methodname{}-C uses the same online PPO training procedure as \methodname{}. This chunked correction makes the learning problem more challenging as the residual policy predicts corrections for future timesteps without access to the intermediate states. See \autoref{app:closed-loop-ablation} for details.

\subsubsection{Real-World Baselines}

We compare the real-world performance of two policies: (1) \textbf{Real-Only}: Diffusion policies trained exclusively on real-world demonstrations $\mathcal{D}_{\text{real}}$, using either 10 or 40 demonstrations, e.g., denoted 10 Real-Only. (2) \textbf{Real+Sim}: Following our sim-to-real approach described in \autoref{sec:sim-to-real-pipeline-method}, we combine the same real-world demonstrations with our synthetic rendered dataset, e.g., denoted 10 Real+Sim.



\section{Experimental Results}
\label{sec:experiments}

Our experimental evaluation focuses on three key aspects. First, we analyze how augmenting chunked Behavior Cloning (BC) policy with closed-loop residual Reinforcement Learning (RL) enables reliable execution of precision-critical manipulation tasks (\autoref{sec:experiments-fine-tuning}). Second, through ablation studies, we identify design choices crucial for \methodname{}'s performance gains (\autoref{sec:resip-ablations}). Finally, we evaluate our method on physical robot hardware (\autoref{sec:experiments-real-world}), demonstrating improved real-world performance through teacher-student distillation while analyzing distillation bottlenecks.

\subsection{Augmenting Trajectory Planning with Reactive Control}
\label{sec:experiments-fine-tuning}

In simulation experiments, we evaluate the fundamental limitations of state-of-the-art BC methods on precision-critical tasks. Using a suite of 6 assembly tasks and 50 demonstrations per task, we find that the Diffusion Policy (DP) architecture struggles with high-precision requirements. Using the \roundtable{} task as an example, DP achieves a 12\% success rate with 50 demonstrations. Increasing the number of demonstrations naturally improves performance. However, as shown in \autoref{fig:scaling-round-table}, scaling to 10,000 demonstrations only improves performance to 56\% on the \roundtable{} task. Another potential solution is addressing the distribution shift through online data collection. While our DAgger implementation shows improved performance over pure BC, it plateaus at 71\% success on the \roundtable{} task, still falling significantly short of expert-level performance (similar scaling behavior is observed for the \oneleg{} task, see \autoref{fig:overview} (Right)).

Our residual learning approach, \methodname{}, addresses these limitations with closed-loop control learned with RL. Starting from the same 50 demonstrations, \methodname{} achieves 94\% on \roundtable{} (up from 12\%) and 98\% success on \oneleg{} (up from 54\%), showing significant gains over the alternatives. The most dramatic improvement is seen in the \peghole{} task, where success rates increase from 5\% to 99\%, highlighting the particular effectiveness of our method when precise local corrections are the primary challenge. \autoref{tab:performance_comparison} shows comparisons across our task suite, demonstrating \methodname{}'s consistent advantages over both BC baselines and alternative RL methods.

\begin{figure}
    \centering
    \includegraphics[width=0.98\linewidth]{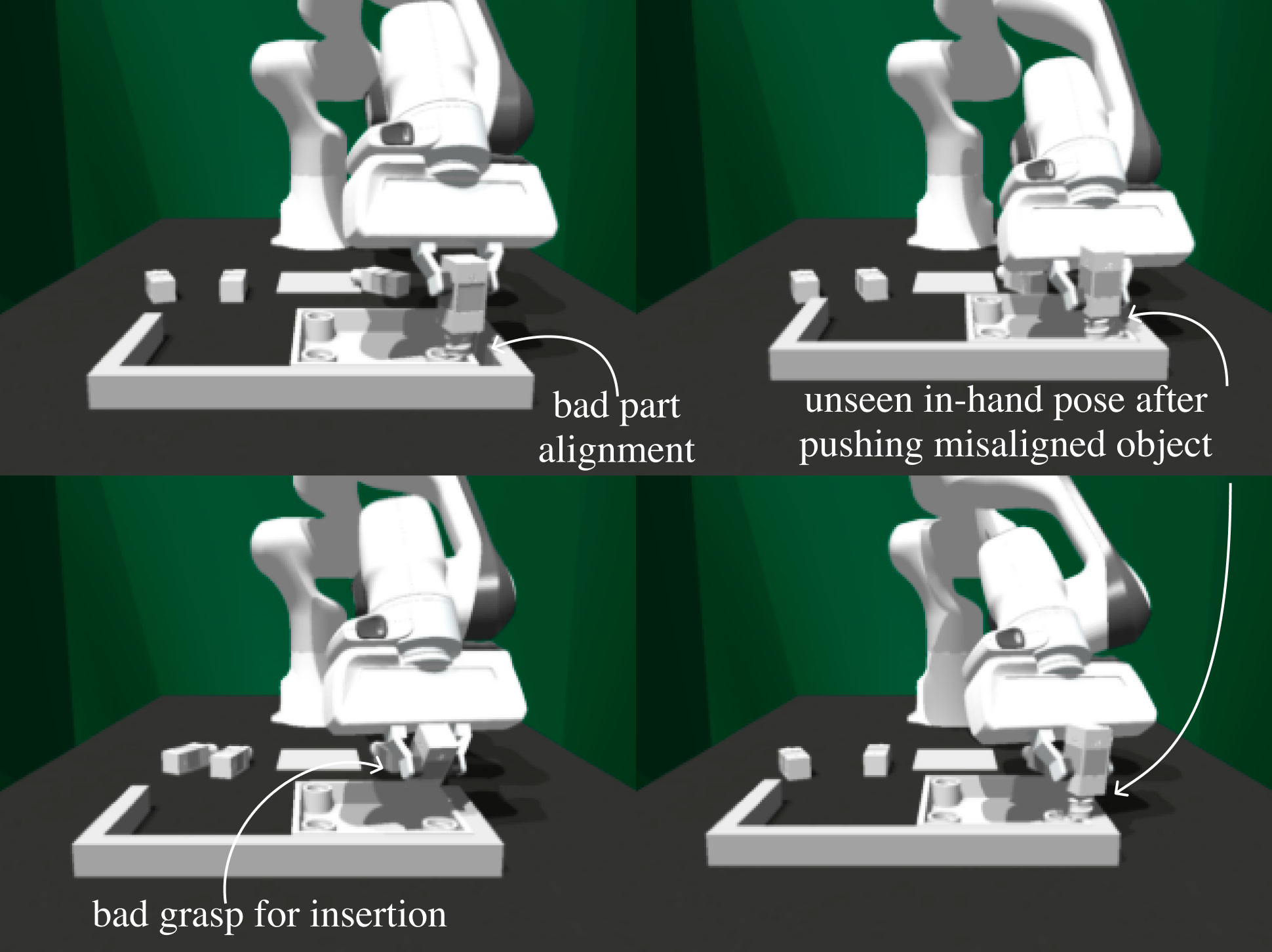}
    \caption{Examples of the common failures of our DP policies in tasks requiring precise part alignments. These failures can be mitigated by small corrections, making our residual framework, \methodname{}, well-suited to improve the reliability of the nominal DP policy.}
    \label{fig:common-errors}
    \vspace{-15pt}
\end{figure}

\subsubsection{Analyzing Failure Modes}

To understand where the large performance improvements stem from, we analyzed failure modes of the DP policies, shown in \autoref{fig:common-errors} and \autoref{fig:overview} (Left). In the low randomization setting, we observe that DP's failures primarily arise from small imprecisions: a common error is pushing the leg down before achieving perfect alignment with the hole. Consequently, the object's pose shifts slightly in the gripper, causing an out-of-distribution grasp pose. The residual policy reliably corrects these errors through minimal adjustments: performing small sideways translations while avoiding premature downward force, only allowing insertion once proper alignment is achieved. We also find that the residual policy makes subtle improvements to initial grasps, enabling more precise downstream alignment between the grasped object and the receptacle.
Based on these observations, we posit that \methodname{}'s significant performance improvements stem from its ability to provide small adjustments in crucial moments---a natural fit for precision assembly tasks where minor misalignments often cause failure. While the base DP policy struggles with precise alignments and insertions, the learned residual component maintains proper alignment through minimal corrections.

However, the performance saturates at 70\%-77\% in the medium randomization settings, with parts initialized up to $\pm$5cm from nominal positions. This performance drop aligns with the intuition of the residual policy as improving precision and reactivity primarily as a local correction mechanism. When parts are placed far from their nominal positions, the DP policy may generate trajectories that deviate too far for the local correction mechanism to correct, leading to failures that are not easily corrected. Even if the failures can be corrected, the resulting states may be out-of-distribution for the DP policy, making corrections infeasible.

\begin{figure}
    \centering
    \includegraphics[width=0.98\linewidth]{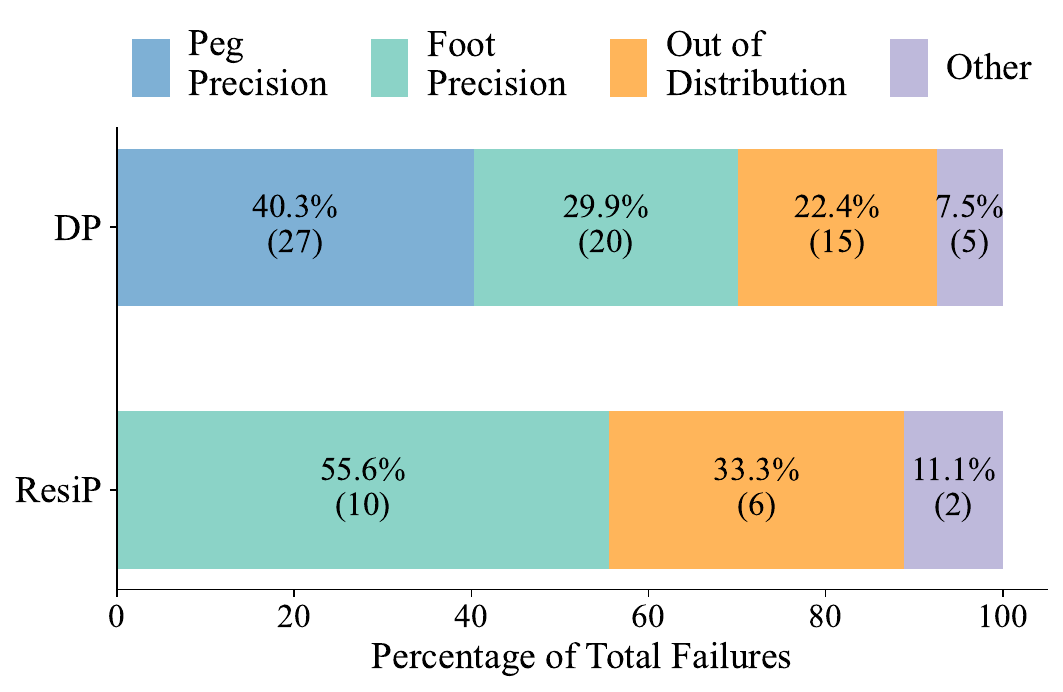}
    \caption{A qualitative analysis of the failure modes between the pre-trained BC policy and the fine-tuned residual policy for the \roundtable{} task on medium randomness. For the pre-trained policy, a large portion of the failures (40\%) relates to the table leg's insertion precision. In the fine-tuned policy, this share dropped to 0\%.}
    \label{fig:round-table-failure-modes}
    \vspace{-15pt}
\end{figure}

To better understand the failure modes of DP and \methodname{} in this higher randomness setting, we manually analyzed 75 randomly sampled trajectories for the \roundtable{} task on medium randomness (see the \website{} for videos). We categorized the observed failures into four primary types:

\begin{enumerate}
    \item `Peg Precision' failures that occur during picking, inserting, or screwing operations with the table leg
    \item `Foot Precision' failures during picking, inserting, or screwing the table foot
    \item `Out-of-Distribution' failures when parts are unreachable by the base policy
    \item `Other' failures are attributed to contact-modeling issues in simulation (e.g., object penetration causing unrealistic accelerations) or timeout conditions.
\end{enumerate}

Our qualitative analysis of these failure modes and recovery behaviors for the \roundtable{} task, shown in \autoref{fig:round-table-failure-modes}, shows that \methodname{} significantly improves reliability during precise insertion phases compared to the baseline approach, and eliminates the precision errors for ``Peg Precision'' entirely but that the `Out-of-Distribution' increases in share.

Surprisingly, \methodname{} also leads to the emergence of qualitatively different behaviors compared to the nominal DP policy. We refer to the \website{}'s uncut videos of rollouts for both policies. In these videos, we observe that \methodname{} has discovered alternative grasping strategies not present in the demonstration data---while demonstrations only showed grasping the table foot from above its center, \methodname{} learned to grasp from the side when the center was unreachable. Additionally, \methodname{} exhibits more decisive recovery behaviors, making deliberate adjustments that lead to successful task completion where the nominal BC policy would ineffectively oscillate back and forth.

\begin{figure}
    \centering
    \includegraphics[width=\linewidth]{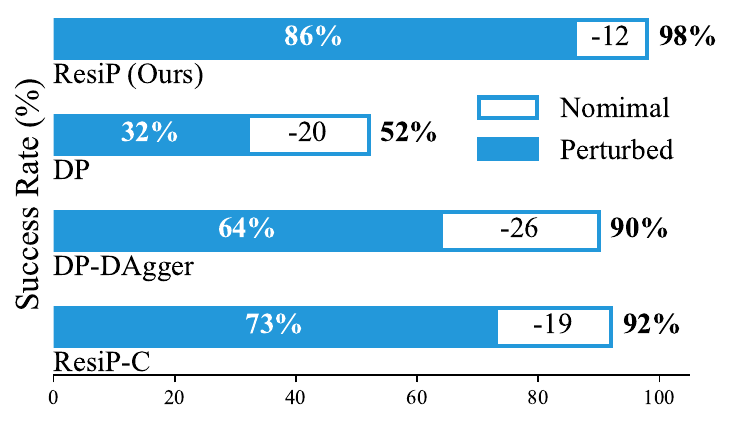}
    \caption{Performance robustness when introducing random force perturbations during evaluation. At each timestep, we randomly apply forces to the objects in the environment, simulating unexpected disturbances. Our method (\methodname{}), which makes per-timestep corrections, shows greater resilience with only a 12\% performance drop compared to nominal conditions. In contrast, chunk-based methods (DP, DP-DAgger, and \methodname{}-C) see larger drops of 19-26\%. This difference in robustness highlights the value of closed-loop control for handling dynamic disturbances not seen during training. All methods were evaluated across 1024 episodes, both with and without perturbations.}
    \label{fig:online_perturbation_comparison}
    \vspace{-15pt}
\end{figure}

\subsection{What drives performance of \methodname{}?}
\label{sec:resip-ablations}

\begin{table*}
    \centering
    \resizebox{\textwidth}{!}{
    \begin{tabular}{lcccccccccc}
    \toprule
    \multirow{2}{*}{Training data} & \multicolumn{2}{c}{Corner} & \multicolumn{2}{c}{Grasp} & \multicolumn{2}{c}{Insert} & \multicolumn{2}{c}{Screw} & \multicolumn{2}{c}{Complete} \\
    \cmidrule(lr){2-3} \cmidrule(lr){4-5} \cmidrule(lr){6-7} \cmidrule(lr){8-9} \cmidrule(lr){10-11}
     & Part & Obs & Part & Obs & Part & Obs & Part & Obs & Part & Obs \\
    \midrule
    10 Real-Only & 5/10 & 5/10 & 5/10 & 7/10 & 2/10 & 3/10 & 0/10 & 2/10 & 0/10 & 2/10 \\
    10 Real+Sim & 9/10 & 9/10 & 7/10 & 8/10 & 0/10 & 3/10 & 0/10 & 3/10 & 0/10 & 3/10 \\
    \addlinespace[.5ex]
    \hdashline
    \addlinespace[1ex]
    40 Real-Only & 10/10 & 8/10 & 9/10 & 8/10 & 6/10 & 3/10 & 2/10 & 3/10 & 2/10 & 3/10\\
    40 Real+Sim & 10/10 & 10/10 & 9/10 & 10/10 & 6/10 & 7/10 & 5/10 & 6/10 & 5/10 & 6/10\\
    \bottomrule
    \end{tabular}
    }
    \caption{Comparing the effect of combining real-world demonstrations with simulation trajectories from our RL-trained residual policies. Co-training with real and synthetic data improves motion quality and success rate on the \oneleg{} task.}
    \label{tab:real-world-real-cotrain}
\end{table*}

\begin{figure*}
    \centering
    \includegraphics[width=0.98\linewidth]{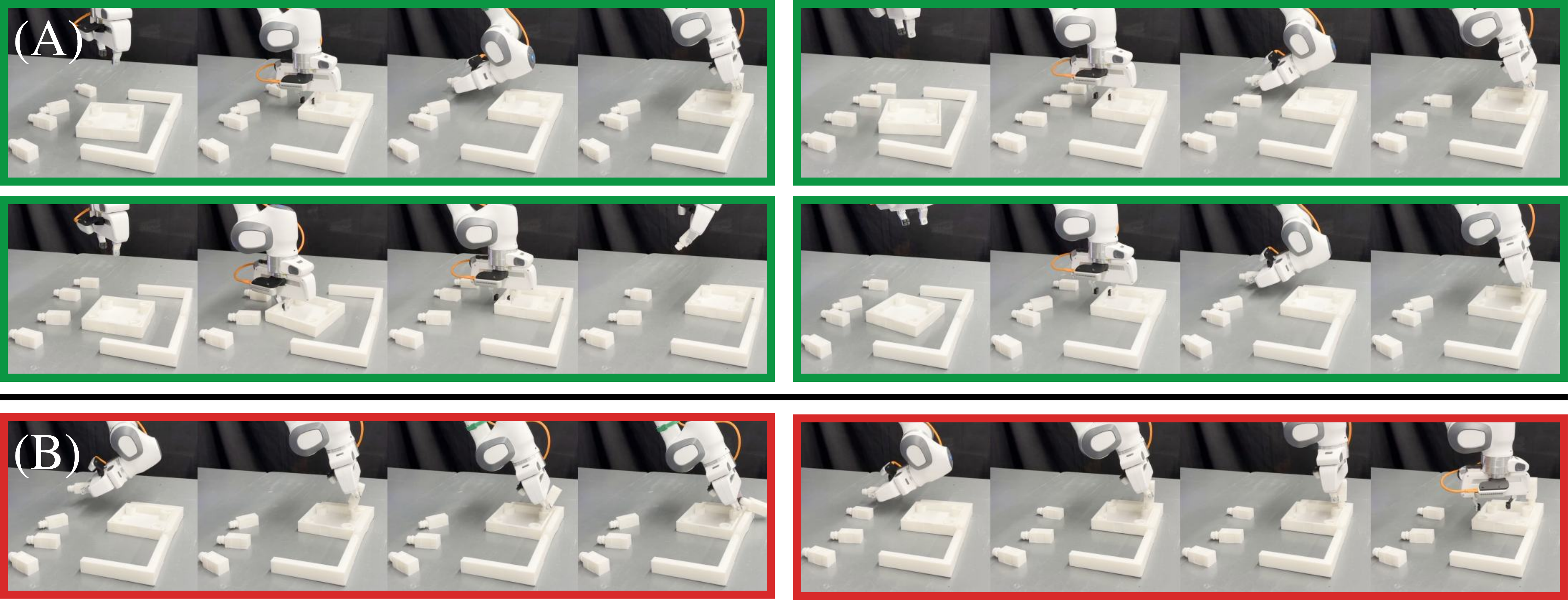}
    \caption{\textbf{(A)} Examples of successful real world assembly from RGB. Co-training with simulation data reduces jerkiness and improves insertion robustness by containing a higher diversity of part poses and insertion locations (see Table~\ref{tab:real-world-real-cotrain}). \textbf{(B)} Example failure: difficulty adjusting the insertion angle/position when grasps lead to unseen in-hand part poses.}
    \label{fig:real-world-sequences}
    \vspace{-15pt}
\end{figure*}

This section investigates different aspects of \methodname{} that improve task performance: (1) training stability and sample efficiency across RL methods, (2) the impact of addressing distribution shift through online data collection, (3) the benefits of closed-loop control compared to chunk-based execution, and (4) robustness to dynamic perturbations.

\subsubsection{Performance and Training Characteristics of RL Methods}

Our residual learning approach, \methodname{}, significantly improves upon baseline methods across all tasks. Using 50 demonstrations per task, where \methodname{} achieved 98\% on the \oneleg{} task (\autoref{tab:performance_comparison}), the alternative RL approaches show more limited improvement of  70\% for PPO-C and 57\% for IDQL. Similar trends hold for all tasks as seen in \autoref{tab:performance_comparison}.

We also see distinct training characteristics across methods. PPO-C, which directly fine-tunes the chunked MLP policy, exhibits training instability and requires careful KL regularization to avoid collapse~\cite{rudner2021pathologies, galashov2019information, schulman2017equivalence}. IDQL faces a different challenge: even with maximum stochasticity in the denoising process, the base DP model produces actions with insufficient variance for effective exploration, limiting the potential for Q-learning-based improvement.

In contrast, \methodname{} demonstrates stable training behavior. Its architecture naturally constrains corrections to be local adjustments to the base policy's absolute pose predictions rather than operating in the full workspace coordinate frame. This local prediction scope, combined with the residual network's small parameter count, prevents the large policy updates that typically destabilize deep RL training~\cite{schulman2015trust, kumar2020conservative}. This stability provides consistent performance across hyperparameters (see \autoref{fig:alpha-ablation} in \appref{app:alpha-ablation}) and enables multiple gradient steps per collected trajectory, improving sample efficiency.

\subsubsection{Impact of Distribution Shift}

To probe the impact of mitigating distribution shift, we compared the baseline DP against DP-DAgger trained using online data collection via DAgger~\cite{ross2011_dagger} as described in \autoref{sec:baselines}. While DP-DAgger significantly outperforms the baseline DP (see yellow vs. red lines in \autoref{fig:overview} (right) and \autoref{fig:scaling-round-table}), it still trails \methodname{} (blue line) by 9\% and 23\% for the \oneleg{} and \roundtable{} tasks, respectively. The remaining performance gap suggests that reducing distribution shift with online data collection does not fully explain \methodname{}'s performance benefits.

\subsubsection{Impact of Closed-Loop Control}
\label{sec:closed-loop-results}

To quantify the benefits of per-timestep corrections, we compared \methodname{} against \methodname{}-C, which predicts corrections at the chunk level as described in \autoref{sec:closed-loop-ablation}. Our experiments reveal that chunk-level corrections lead to significantly slower learning progress, with \methodname{}-C requiring significantly more environment interactions to achieve comparable performance (\autoref{fig:chunked-residual-ablation} in \appref{app:closed-loop-ablation}). Moreover, even with extended training, \methodname{}-C's performance saturates at 92\% compared to \methodname{}'s 98\% on the \oneleg{} task. This performance gap suggests that the ability to make frequent corrections accelerates learning and enables higher terminal performance.

\subsubsection{Robustness to Dynamic Perturbations}

To further evaluate the benefits of closed-loop control, we compare the performance degradation of different methods under dynamic disturbances as described in \autoref{sec:evaluation-protocol}. As shown in \autoref{fig:online_perturbation_comparison}, methods using chunk-based execution---DP, DP-DAgger, and \methodname{}-C---experience significant performance drops of 19-26 percentage points under these perturbations. In contrast, \methodname{}, with its ability to make per-timestep corrections, maintains a more robust performance with a 12\% drop. This significant difference in robustness (7-14 percentage points) provides further evidence for the value of closed-loop control in precise manipulation tasks.


\subsection{Real-World Deployment}
\label{sec:experiments-real-world}

\begin{figure}
    \centering
    \includegraphics[width=0.98\linewidth]{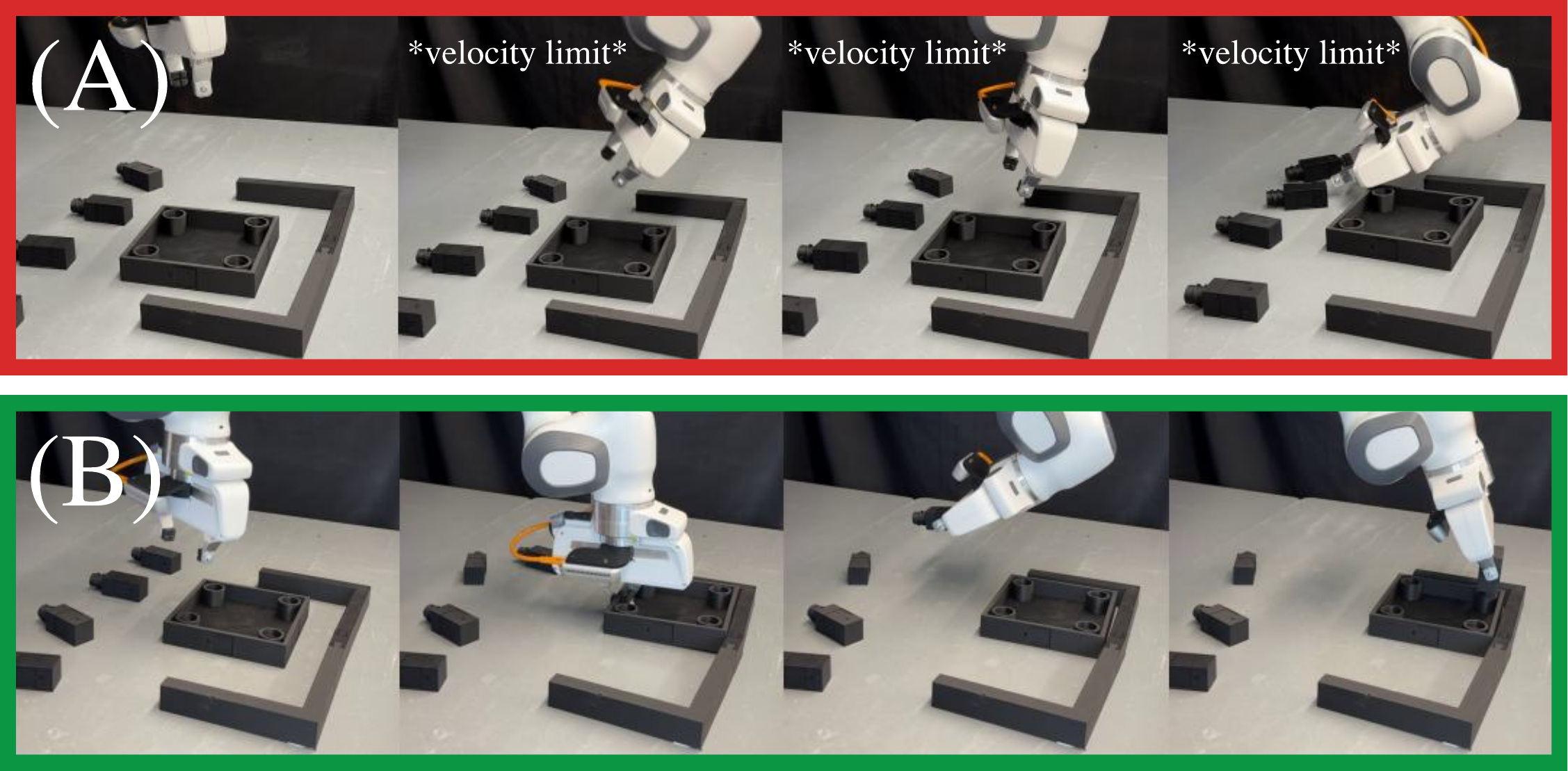}
    \caption{\textbf{(A)} When changing the part appearances from white to black, the policy trained on real data only (Real-Only) seizes to function. The behavior becomes erratic, and all trials ended with the robot hitting a velocity limit. \textbf{(B)} When mixing in synthetic data with more diverse colors (see \autoref{fig:synthetic-data}), the policy (Real+Sim) regains the ability to complete the task, though with lower performance than for white parts.}
    \label{fig:real-black-parts}
    \vspace{-15pt}
\end{figure}

\subsubsection{Real-World Performance}

In our sim-to-real experiments, following the protocol described in \autoref{sec:evaluation-protocol}, we find that the Real+Sim policies achieve significantly higher success rates (50-60\%) compared to Real-Only baselines (20-30\%) on the \oneleg{} task (\autoref{tab:real-world-real-cotrain}). Qualitatively, the co-trained policy (Real+Sim) exhibits smoother behavior with fewer erratic movements that exceed robot hardware limits. \autoref{fig:real-world-sequences} shows typical behaviors: Row (A) shows successful assembly sequences, while Row (B) demonstrates the primary failure mode---imprecise alignment before release, mirroring challenges observed in simulation.

To assess robustness to visual variations, we tested changing part colors from the white used in data collection to an unseen black. Unsurprisingly, the Real-Only policy fails catastrophically, triggering velocity limits on every trial, as shown in \autoref{fig:real-black-parts} (A). In contrast, Real+Sim, trained with the same real-world data but with color-randomized synthetic data (see \autoref{fig:synthetic-data}; bottom), maintains basic functionality though with reduced performance compared to the original color scheme. While highlighting the benefits of synthetic data, this also indicates opportunities for improving sim-to-real transfer. See \appref{app:experiments-real-world} for more detailed analysis.

\subsubsection{Understanding Performance Limitations}

\begin{figure}
    \centering
    \includegraphics[width=\linewidth]{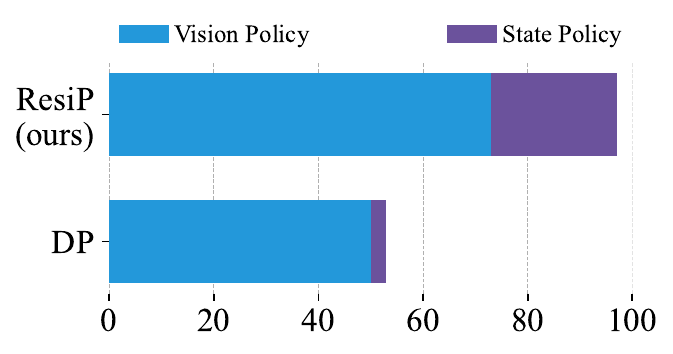}
    \caption{A comparison of performance between state and vision policies in policy distillation. The Diffusion Policy (DP) baseline achieves similar performance with both state and vision inputs when trained directly on the same 50 demonstrations (left), indicating that the vision modality does not inherently limit performance. However, we observe a significant performance gap when distilling our state-based \methodname{} policy into a vision-based policy (right). This consistent performance gap suggests the challenge lies in the distillation process rather than the vision modality, as direct vision-based training shows no such degradation.}
    \label{fig:bc_vs_rl_distillation}
    \vspace{-15pt}
\end{figure}

To understand the gap between simulation and real-world performance, we hypothesize three potential limiting factors: (1) the change from state- to vision-based observations, (2) the sim-to-real gap, and (3) the policy distillation process. Comparative experiments between image and state-based students show similar performance gaps from the teacher's 98\% success rate (\autoref{fig:bc_vs_rl_distillation}), ruling out the change in observation modality as the primary bottleneck. Furthermore, our scaling analysis shows that even increasing synthetic trajectories from 10k to 100k only marginally improves success rates from 78\% to 80\% (\autoref{fig:overview}; Right), indicating a fundamental limitation in the policy distillation process itself rather than purely sim-to-real challenges. These findings motivate future investigations into better approaches for vision-based policy distillation, particularly focusing on online learning and closed-loop control. See \appref{app:distillation-analysis} for detailed analysis.

\section{Related Works}
\label{sec:related}

\subsection{Training diffusion models with reinforcement learning}
A fundamental challenge in applying RL to diffusion models is that the final action probabilities are not directly accessible due to the iterative nature of the denoising process, making policy gradient methods difficult to apply. Recent work has explored various approaches to combining diffusion models with RL~\cite{fan2023optimizing, fan2023dpok, black2023training, li2024learning_diffusion_pg, ren2024diffusion}. Some approaches cast diffusion de-noising as a Markov Decision Process~\cite{black2023training, fan2023dpok}, enabling preference-aligned image generation with policy gradient RL, but suffer from training instability. While \cite{ren2024diffusion} introduced more stable direct diffusion policy fine-tuning, their method remains architecture-specific and lacks closed-loop control.

Other approaches include Q-function-based importance sampling~\cite{hansen-estruch_idql_2023}, advantage weighted regression~\cite{goo2022know}, and return-conditioned supervised learning~\cite{chen2021decision,ajay_is_2023, janner_planning_2022}. Some methods augment the denoising objective with Q-function maximization~\cite{wang2022diffusion} or iteratively update the dataset using Q-functions~\cite{yang2023policy}. However, these approaches mainly enable better \textit{extraction} or stitching of existing behaviors rather than learning new, \textit{corrective} behaviors. Recent work on training diffusion policies from scratch~\cite{li2024learning_diffusion_pg} uses complex mechanisms like unsupervised clustering and Q-learning ensembles for multi-modal behavior discovery.

Beyond the iterative denoising challenge, modern BC approaches introduce additional complications through action chunking, where policies predict sequences of multiple future actions. While this improves BC performance, it creates significant challenges for RL fine-tuning by expanding the action space---for instance, chunks of 8 actions result in an 8-fold increase in action dimensionality. This issue affects most modern BC architectures like ACT~\cite{zhao_rss23_aloha, drolet2024comparison}, as they rely on action chunking. Policy gradient methods struggle with such high-dimensional action spaces, particularly when applied to deep neural networks~\cite{hu_imitation_2023_bootstrapped}. Furthermore, recent work shows that RL fine-tuning of large pre-trained models can lead to forgetting of pre-training capabilities~\cite{wolczyk2024fine}. Our method avoids these problems by keeping the base policy frozen and training only a small residual model, which preserves the pre-trained capabilities and enables stable policy gradient training with closed-loop control.

\subsection{Residual learning in robotics}

Learning corrective residual components in conjunction with learned or non-learned ``base'' models has been widely successful in robotics.
Common frameworks include learning residual policies that correct for errors made by a nominal behavior policy~\cite{silver2018residual, johannink2019residual, alakuijala2021residual, davchev2022residual, carvalho2022residual, shafiullah2022behavior_bet, haldar2023teach_fish_residual, lee2024behavior_vqbet} and combining learned components to correct for inaccuracies in analytical models for physical dynamics~\cite{ajay2018augmenting, kloss2022combining, zeng2020tossingbot} or sensor observations~\cite{kaufmann2023champion_residual}. Unlike prior residual policy learning approaches, our method uniquely combines RL-based training with running the base at a lower frequency, with the residual providing corrections at every control step.

Residual policies have been used in insertion applications~\cite{schoettler2020deep_residual_insertion}, and recent work has applied residual policy learning to the same FurnitureBench task suite we study in this paper~\cite{jiang2024transic_assembly}. Their approach uses the residual component to model online human-provided corrections via supervised learning, whereas we train our residual policy from scratch with RL using task rewards in simulation.



\section{Discussion}

\balance

The local nature of our residual policies is designed to complement rather than replace the trajectory planning capabilities of the base policy. While this design enables precise corrections, our approach still relies on the base policy for macro-level behaviors.
As such, our proposed method struggles in regimes with very high initial scene randomness, as both the base policies and actions produced via RL exploration struggle to deal with out-of-support initial part poses.
Our imitation learning scaling analyses were conducted using a dataset from an RL expert, not human demonstrations. These two demonstration sources will likely have different distributions, which may change the analyses. However, acquiring 100k demonstrations from a human demonstrator was not feasible in the present work.
Furthermore, despite showcasing the advantage of incorporating simulation data, sim-to-real for vision-based policies still presents a challenge. There remains a performance gap in both teacher-student distillation and sim-to-real distribution shifts.
Future investigations may include better sim-to-real transfer techniques, exploration mechanisms for discovering how to correct large-scale execution errors, tractable interactive learning for real-world policy distillation, and incorporating inductive biases (like~\cite{simeonov2022neural}) that help generalize to broader initial state distributions.

\section*{Acknowledgments}

This work was partly supported by the Sony Research Award, the US Government, and the Hyundai Motor Company. The computations in this paper were run on the FASRC cluster, supported by the FAS Division of Science Research Computing Group at Harvard University, and on the MIT Supercloud\cite{reuther_interactive_2018}. Experiment tracking and model checkpoint storage were provided by \href{https://wandb.ai/}{Weights and Biases}.

\subsection*{Author Contributions}

\textbf{LA} led the project and implemented most of the code and training infrastructure, including the main residual PPO implementation, and is the primary author.

\textbf{AS} helped conceive of and advise on project goals, led deployment on real hardware and sim-to-real rendering, and helped write the paper.

\textbf{IS} led the implementation of reinforcement learning baselines, helped debug residual PPO implementation, and helped write the paper.

\textbf{MT} provided valuable insights, recommendations, and discussions
on reinforcement learning fine-tuning and sim-to-real transfer.

\textbf{PA} advised the project and provided valuable feedback on project framing, contributions, and writing.




\clearpage


\bibliographystyle{IEEEtran}
\bibliography{IEEEabrv,references}{}

\begin{appendices}
\onecolumn




\section{Implementation Details}
\label{sec:implementation-details}

\subsection{Training Hyperparameters}
\label{app:hyperparameters}

\subsubsection{State-based behavior cloning}

We provide a detailed set of hyperparameters used for training. General hyperparameters for all models can be found in \autoref{tab:training-hyperparameters}, while specific hyperparameters for the diffusion models are in \autoref{tab:diffusion-model-hyperparameters}, and those for the MLP baseline are in \autoref{tab:mlp-baseline-hyperparameters}.

\begin{table}[H]
\centering
\caption{Training hyperparameters shared for all state-based BC models}
\label{tab:training-hyperparameters}
\begin{tabular}{p{0.4\linewidth}p{0.3\linewidth}}
\toprule
Parameter & Value \\
\midrule
Control mode & Absolute end-effector pose \\
Action space dimension & 10 \\
Proprioceptive state dimension & 16 \\
Orientation Representation & 6D~\cite{zhou2019continuity} \\
Max LR & $10^{-4}$ \\
LR Scheduler & Cosine \\
Warmup steps & 500 \\
Weight Decay & $10^{-6}$ \\
Batch Size & 256 \\
Max gradient steps & 400k \\
\bottomrule
\end{tabular}
\end{table}

\begin{table}[H]
\centering
\caption{State-based diffusion pre-training hyperparameters}
\label{tab:diffusion-model-hyperparameters}
\begin{tabular}{p{0.4\linewidth}p{0.3\linewidth}}
\toprule
Parameter & Value \\
\midrule
U-Net Down dims & $[256, 512, 1024]$ \\
Diffusion step embed dim & $256$ \\
Kernel size & $5$ \\
N groups & $8$ \\
Parameter count & 66M \\
Observation Horizon $T_o$ & 1 \\
Prediction Horizon $T_p$ & 32 \\
Action Horizon $T_a$ & 8 \\
DDPM Training Steps & 100 \\
DDIM Inference Steps & 4 \\
\bottomrule
\end{tabular}
\end{table}

\begin{table}[H]
\centering
\caption{State-based MLP pre-training hyperparameters}
\label{tab:mlp-baseline-hyperparameters}
\begin{tabular}{p{0.4\linewidth}p{0.3\linewidth}}
\toprule
Parameter & Value \\
\midrule
Residual Blocks & 5 \\
Residual Block Width & 1024 \\
Layers per block & 2 \\
Parameter count & 11M \\
Observation Horizon $T_o$ & 1 \\
Prediction Horizon $T_p$ (S / C) & 1 / 8 \\
Action Horizon $T_a$ (S / C) & 1 / 8 \\
\bottomrule
\end{tabular}
\end{table}

\subsubsection{State-based reinforcement learning} Below, we list the hyperparameters used for online reinforcement learning fine-tuning. The parameters that all state-based RL methods methods shared are in \autoref{tab:fine-tuning-shared-hyperparameters}. Method-specific hyperparameters for training the different methods are in the tables below, direct fine-tuning of the MLP in \autoref{tab:mlp-finetuning-hyperparameters}, online IDQL in \autoref{tab:online-idql-hyperparameters}, and the residual policy in \autoref{tab:residual-ppo-hyperparameters}. The different methods were tuned independently, but the same hyperparameters were used for all tasks within each method.

\begin{table}[H]
\centering
\caption{Hyperparameters shared for all online fine-tuning approaches}
\label{tab:fine-tuning-shared-hyperparameters}
\begin{tabular}{p{0.4\linewidth}p{0.3\linewidth}}
\toprule
Parameter & Value \\
\midrule
Control mode & Absolute end-effector pose \\
Action space dimension & 10 \\
Proprioceptive state dimension & 16 \\
Orientation Representation & 6D~\cite{zhou2019continuity} \\
Num parallel environments & 1024 \\
Max environment steps & 500M \\
Critic hidden size & 256 \\
Critic hidden layers & 2 \\
Critic activation & ReLU \\
Critic last layer activation & Linear \\
Critic last layer bias initialization & 0.25 \\
Discount factor & 0.999 \\
GAE~\cite{schulman2015high} lambda & 0.95 \\
Clip $\epsilon$ & 0.2 \\
Max gradient norm & 1.0 \\
Target KL & 0.1 \\
Num mini-batches & 1 \\
Episode length, \oneleg{} & 700 \\
Episode length, \lamp{}/\roundtable{} & 1000 \\
Normalize advantage & true \\
\bottomrule
\end{tabular}
\end{table}

\begin{table}[H]
\centering
\caption{Hyperparameters for direct fine-tuning of MLP}
\label{tab:mlp-finetuning-hyperparameters}
\begin{tabular}{p{0.4\linewidth}p{0.3\linewidth}}
\toprule
Parameter & Value \\
\midrule
Update epochs & 1 \\
Learning rate actor & $10^{-4}$ \\
Learning rate critic & $10^{-4}$ \\
Value function loss coefficient & 1.0 \\
KL regularization coefficient & 0.5 \\
Actor Gaussian initial log st.dev. & -4.0 \\
\bottomrule
\end{tabular}
\end{table}

\begin{table}[H]
\centering
\caption{Hyperparameters for training value-augmented diffusion sampling (IDQL)}
\label{tab:online-idql-hyperparameters}
\begin{tabular}{p{0.4\linewidth}p{0.3\linewidth}}
\toprule
Parameter & Value \\
\midrule
Update epochs & 10 \\
Learning rate Q-function & $10^{-4}$ \\
Learning rate scheduler & Cosine \\
Num action samples & 20 \\
Actor added Gaussian noise, log st.dev. & $-4$ \\
\bottomrule
\end{tabular}
\end{table}

\begin{table}[H]
\centering
\caption{Hyperparameters for residual PPO training}
\label{tab:residual-ppo-hyperparameters}
\begin{tabular}{p{0.4\linewidth}p{0.3\linewidth}}
\toprule
Parameter & Value \\
\midrule
Residual action scaling factor & 0.1 \\
Update epochs & 50 \\
Learning rate actor & $3\cdot 10^{-4}$ \\
Learning rate critic & $5\cdot 10^{-3}$ \\
Learning rate scheduler & Cosine \\
Value function loss coefficient & 1.0 \\
Actor Gaussian initial log st.dev. & -1.0 \\
\bottomrule
\end{tabular}
\end{table}

\subsubsection{Image-based real-world distillation}
\label{app:rgb-policy-implementation-details}

We use a separate set of hyperparameters for real-world experiments, presented in \autoref{tab:vision-hyperparameters}. The main difference is that we found in experimentation that the transformer backbone in ~\cite{chi_diffusion_2023} worked better than the UNet for real-world experiments. These models are also operating from RGB observations instead of privileged states, and we provide parameters for the image augmentations applied to the front camera in \autoref{tab:front-hyperparameters} and the wrist camera in \autoref{tab:wrist-hyperparameters}.

\begin{table}[H]
\centering
\caption{Training hyperparameters for real-world distilled policies}
\label{tab:vision-hyperparameters}
\begin{tabular}{p{0.35\linewidth}p{0.35\linewidth}}
\toprule
Parameter & Value \\
\midrule
Control mode & Absolute end-effector pose \\
Action space dimension & 10 \\
Proprioceptive state dimension & 16 \\
Orientation Representation & 6D~\cite{zhou2019continuity} \\
Max policy LR & $10^{-4}$ \\
Max encoder LR & $10^{-5}$ \\
LR Scheduler (both) & Cosine \\
Policy scheduler warmup steps & 1000 \\
Policy scheduler warmup steps & 5000 \\
Weight decay & $10^{-3}$ \\
Batch size & 256 \\
Max gradient steps & 500k \\
Image size input & $2 \times 320 \times 240 \times 3$ \\
Image size encoder & $2 \times 224 \times 224 \times 3$ \\
Vision Encoder Model & ResNet18~\cite{he2016deep} \\
Encoder Weights & R3M~\cite{nair_r3m_2022} \\
Encoder Parameters & $2 \times 11$ million \\
Encoder Projection Dim & 128 \\
Diffusion backbone architecture & Transformer (similar to~\cite{chi_diffusion_2023}) \\
Transformer num layers & 8 \\
Transformer num heads & 4 \\
Transformer embedding dim & 256 \\
Transformer embedding dropout & 0.0 \\
Transformer attention dropout & 0.3 \\
Transformer causal attention & true \\
\bottomrule
\end{tabular}
\end{table}

\begin{table}[H]
\centering
\caption{Parameters for front camera image augmentation}
\label{tab:front-hyperparameters}
\begin{tabular}{p{0.35\linewidth}p{0.35\linewidth}}
\toprule
Parameter & Value \\
\midrule
Color jitter (all parameters) & $0.3$ \\
Gaussian blur, kernel size & $5$ \\
Gaussian blur, sigma & $(0.01, 1.2)$ \\
Random crop area & $280\times 240$ \\
Random crop size & $224\times 224$ \\
Random erasing, fill value & random \\
Random erasing, probability & $0.2$ \\
Random erasing, scale & $(0.02, 0.33)$ \\
Random erasing, ratio & $(0.3, 3.3)$ \\
\bottomrule
\end{tabular}
\end{table}

\begin{table}[H]
\centering
\caption{Parameters for wrist camera image augmentation}
\label{tab:wrist-hyperparameters}
\begin{tabular}{p{0.35\linewidth}p{0.35\linewidth}}
\toprule
Parameter & Value \\
\midrule
Color jitter (all parameters) & $0.3$ \\
Gaussian blur, kernel size & $5$ \\
Gaussian blur, sigma & $(0.01, 1.2)$ \\
Random crop & Not used \\
Image resize & $320\times 240 \rightarrow 224\times 224$ \\
\bottomrule
\end{tabular}
\end{table}

\subsection{Action and State-Space Representations}

\paragraph{Action space} The policies predict 10-dimensional actions consisting of absolute poses in the robot base frame as the actions and a gripper action. In particular, the first 3 dimensions predict the desired end-effector position in the workspace, the next 6 predict the desired orientation using a 6-dimensional representation described below. The final dimension is a gripper action, 1 to command closing gripper and -1 for opening.

\paragraph{Proprioceptive state space} The policy receives a 16-dimensional vector containing the current end-effector state and gripper width. In particular, the first 3 dimensions is the current position in the workspace, the next 6 the current orientation in the base frame (the same 6D representation), the next 3 the current positional velocity, the next 3 the current roll, pitch, and yaw angular velocity, and finally the current gripper width.

\paragraph{Rotation representation}

We use a 6D representation to represent all orientations and rotations for the predicted action, and proprioceptive end-effector pose orientation \cite{zhou2019continuity,levinson2020analysis}.
The poses of the parts in state-based environments are represented with unit quaternions.
While this representation contains redundant dimensions, it is continuous, meaning that small changes in orientation lead to small changes in the representation values, which can make learning easier\cite{zhou2019continuity,levinson2020analysis,geist2024learning}. This is not generally the case for Euler angles and quaternions. The 6D representation is constructed by taking two arbitrary 3D vectors and performing Gram-Schmidt orthogonalization to obtain a third orthogonal vector to the first two. The resulting three orthogonal vectors form a rotation matrix that represents the orientation.
The end-effector rotation angular velocity is still encoded as roll, pitch, and yaw values.

\paragraph{Action and state-space normalization}

All dimensions of the action, proprioceptive state, and parts pose (for state-based environments), were independently scaled to the range [-1, 1]. That is, we did not handle orientation representations (quaternions/6D~\cite{zhou2019continuity}) in any particular way.
The normalization limits were calculated over the dataset at the start of behavior cloning training. They were stored in the actor with the weights and reused as the normalization limits when training with reinforcement learning.
The normalization used here follows the same approach as in previous works such as \cite{reuss_goal-conditioned_2023, chi_diffusion_2023}. This normalization method is widely accepted for diffusion models. In \cite{reuss_goal-conditioned_2023}, the input was standardized to have a mean of 0 and a standard deviation of 1, instead of using min-max scaling to the range of [0, 1]. This approach was not tested in our experiments.

\subsection{Image Augmentation}

During training, we apply image augmentation and random cropping to both camera views. Specifically, only the front camera view undergoes random cropping. We also apply color jitter with a hue, contrast, brightness, and saturation set to 0.3. Additionally, we apply Gaussian blur with a kernel size of 5 and sigma between 0.1 and 5 to both camera views.

At inference time, we statically center-crop the front camera image from 320x240 to 224x224 and resize the wrist camera view to the same dimensions. For both the random and center crops, we resized the image to 280x240 to ensure that essential parts of the scene are not cropped out due to excessive movement.

The values mentioned above were chosen based on visual assessment to balance creating adversarial scenarios and keeping essential features discernible. We have included examples of these augmentations below.

\begin{figure}[H]
    \centering
    \includegraphics[width=0.49\textwidth]{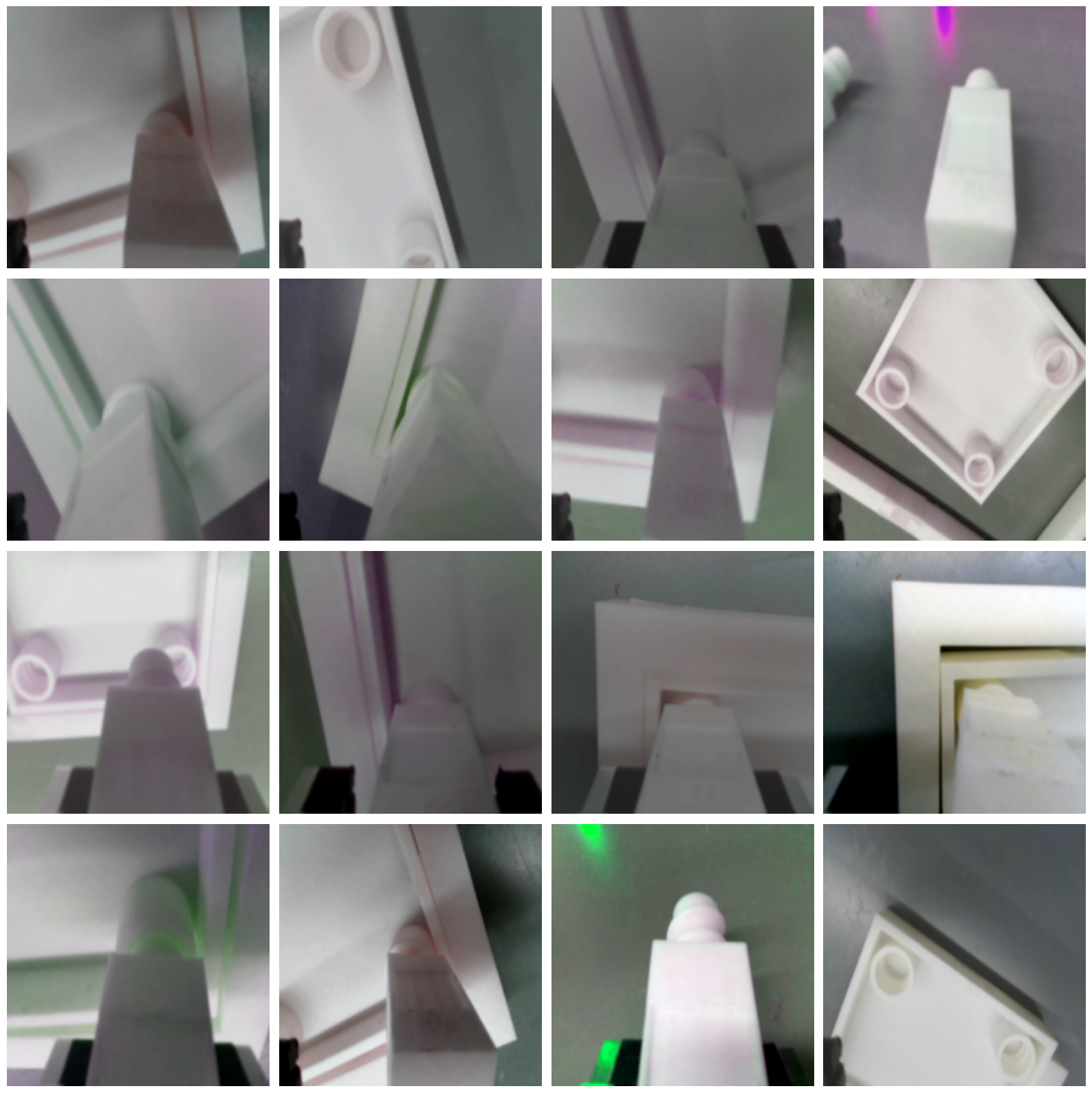}
    \hfill
    \includegraphics[width=0.49\textwidth]{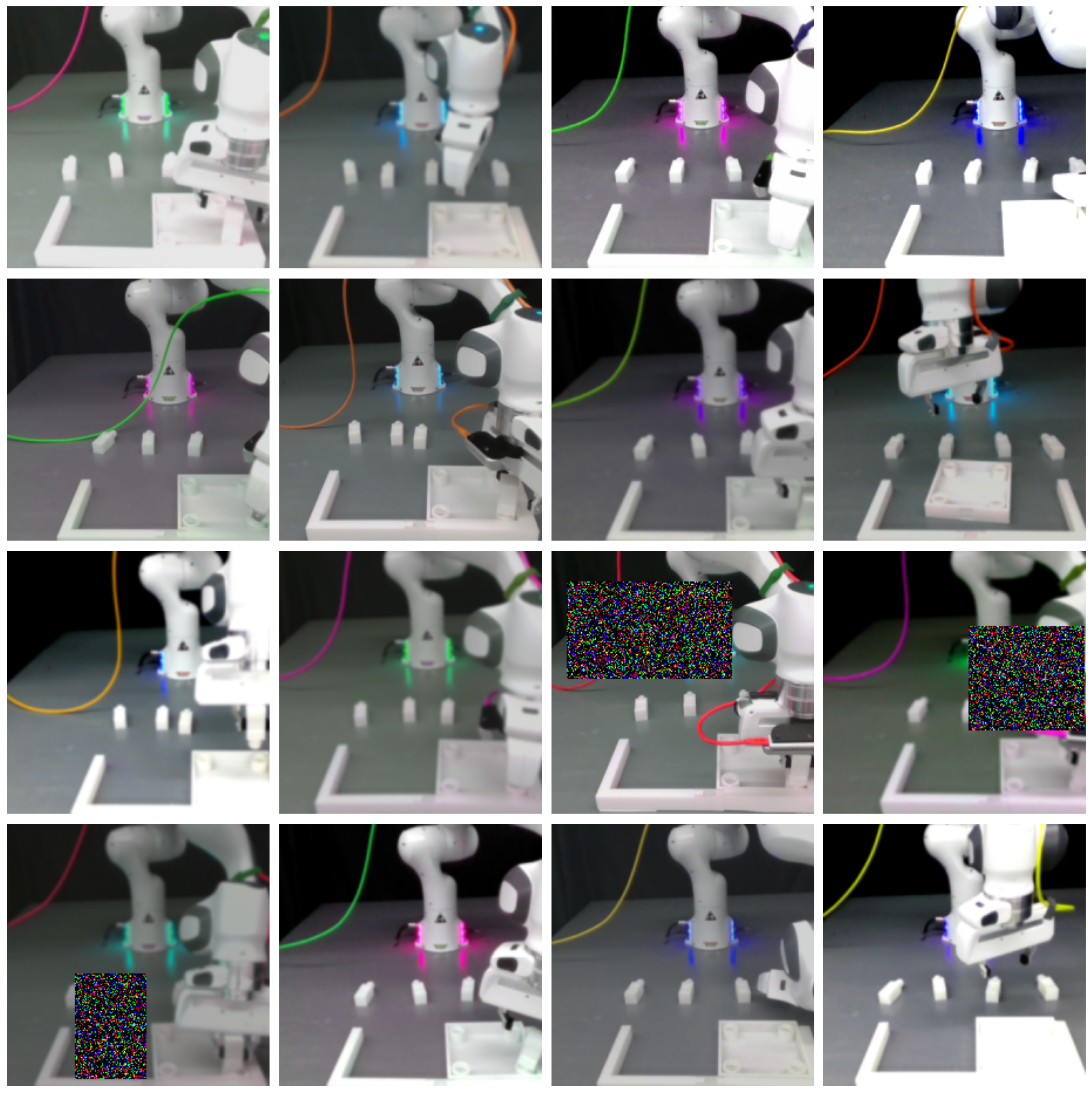}
    \caption{\textbf{Left:} Examples of augmentations of the wrist camera view, consisting of color jitter and Gaussian blur. \textbf{Right:} Examples of augmentations for the front view also consist of color jitter and Gaussian blur augmentations and random cropping.}
    \label{fig:image-aug-example-side-by-side}
\end{figure}

\section{Tasks and Environment}
\label{app:tasks-and-environment}

\subsection{Tasks details and reward signal}
\label{app:tasks-details}

\subsubsection{Furniture assembly tasks}

We detail a handful of differentiating properties for each of the three tasks we use in \autoref{tab:task-attributes}. \oneleg{} involves assembling 2 parts, the tabletop and one of the 4 table legs. The assembly is successful if the relative poses between the parts are close to a predefined assembled relative pose. When this pose is achieved, the environment returns a reward of 1. That is, for the \oneleg{} task, the policy received a reward of 1 only at the very end of the episode. For \roundtable{} and \lamp{}, which consists of assembling 3 parts together, the policy receives a reward signal of 1 for each pair of assembled parts. E.g. for the \lamp{} task, when the bulb is fully screwed into the base, the first reward of 1 is received, and the second is received when the shade is correctly placed.

\subsubsection{Real-to-sim task: \texttt{mug-rack}}
\label{app:mug-rack}
\begin{figure}[H]
    \centering
    \begin{subfigure}[b]{0.3\textwidth}
        \includegraphics[width=\textwidth]{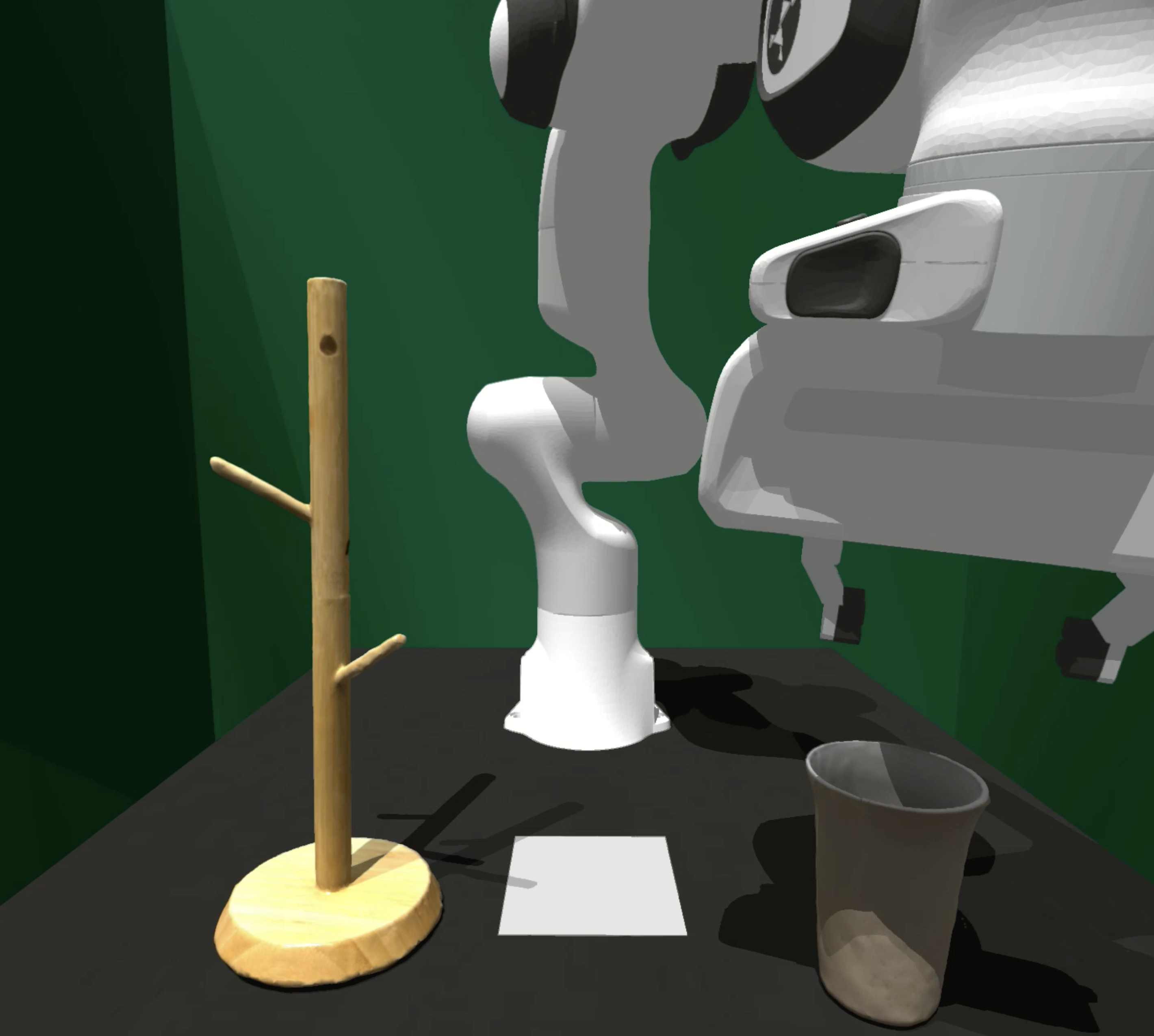}
        \caption{Example task initialization of the \texttt{mug-rack} task.}
    \end{subfigure}
    \hfill
    \begin{subfigure}[b]{0.3\textwidth}
        \includegraphics[width=\textwidth]{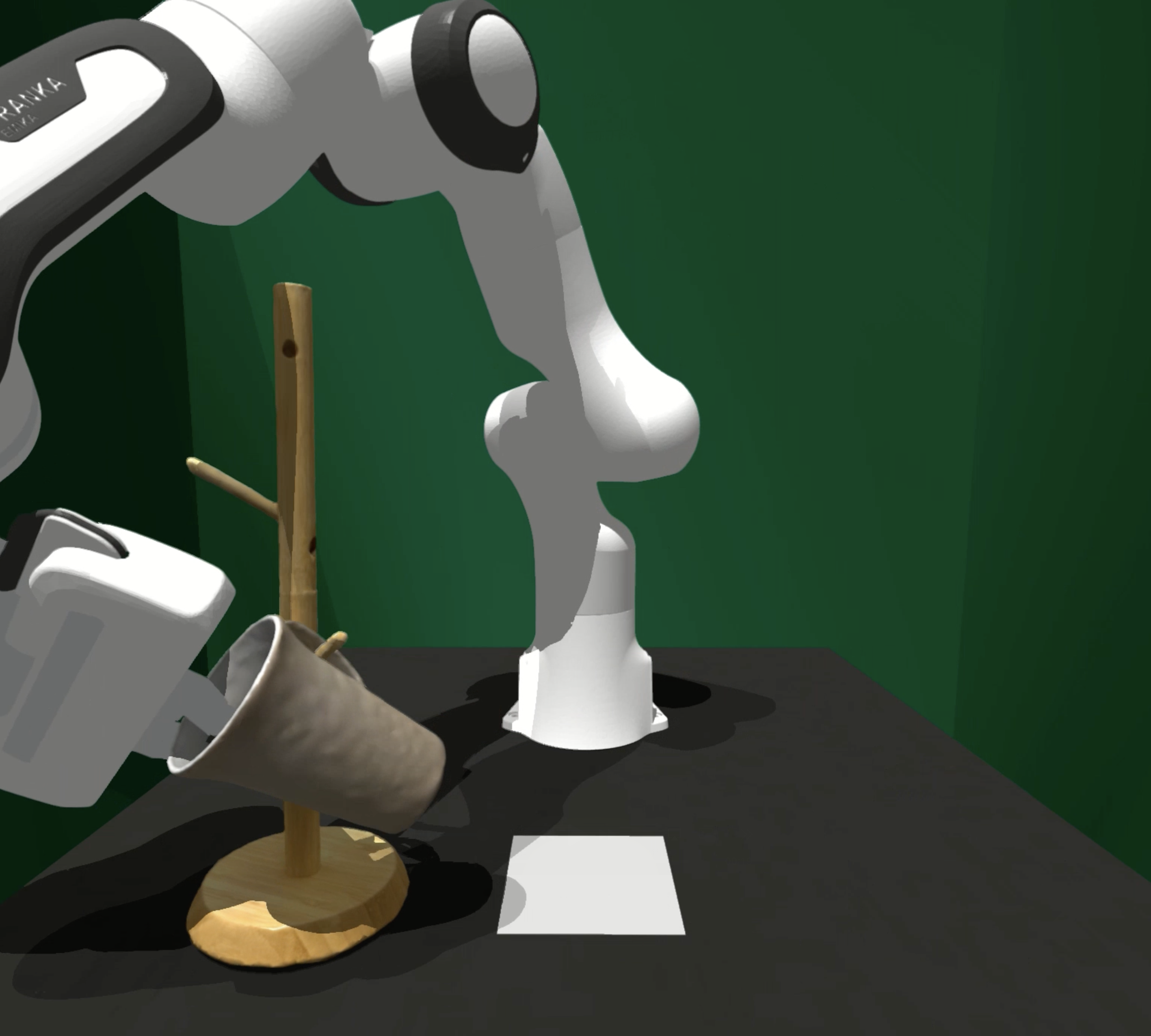}
        \caption{Example of hanging the mug on the lower rack.}
    \end{subfigure}
    \hfill
    \begin{subfigure}[b]{0.3\textwidth}
        \includegraphics[width=\textwidth]{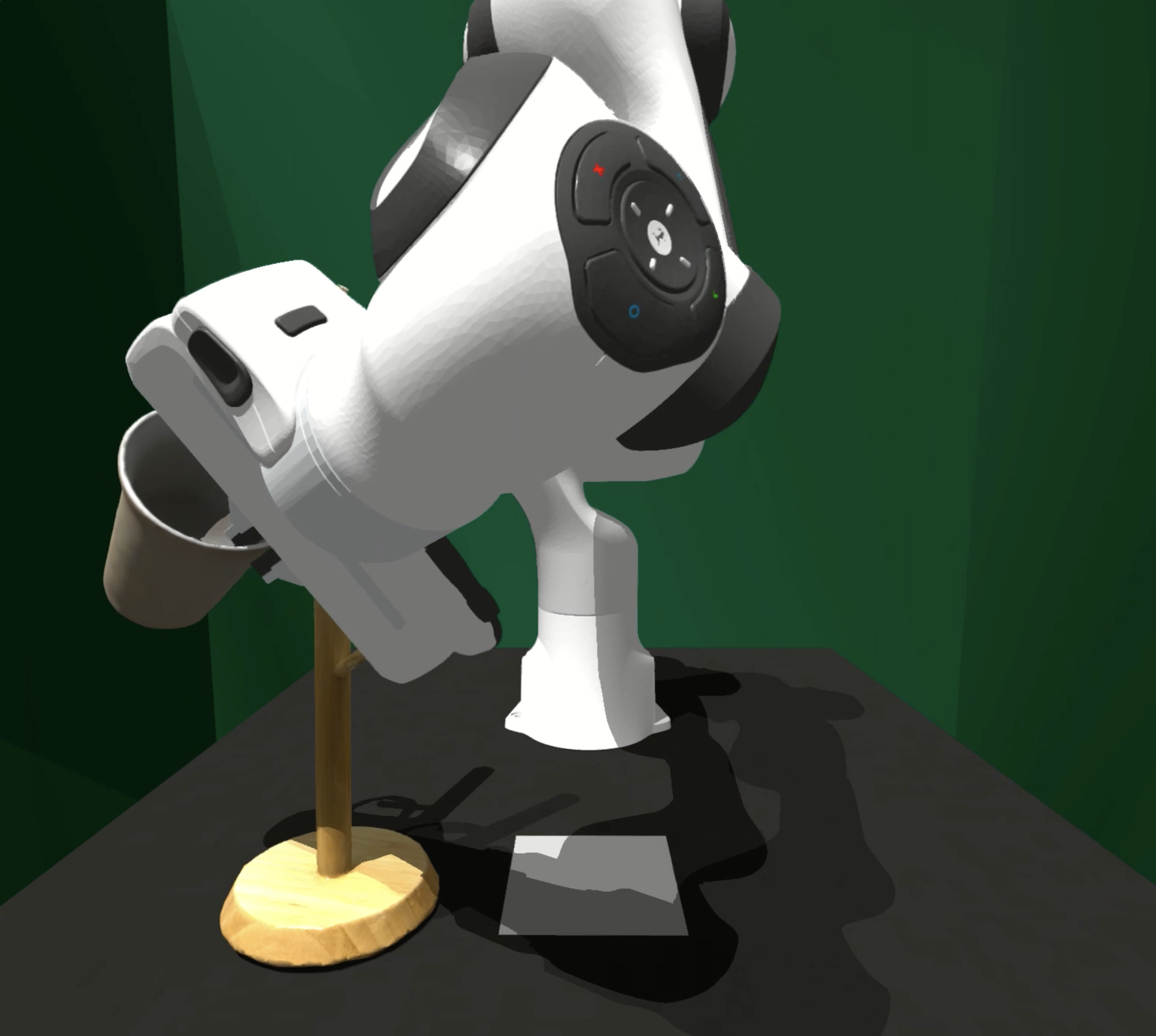}
        \caption{Example of hanging the mug on the upper rack.}
    \end{subfigure}
    \caption{Overview of the \texttt{mug-rack} task to showcase the real-to-sim capabilities one can leverage with our pipeline. This also shows how reward signals can be inferred directly from data instead of being hand-designed. Finally, as the task can be completed in one of several ways, this task also tests the policies' capability to deal with multi-modality.}
    \label{fig:mug-rack}
\end{figure}

This task involves the robot picking up a coffee mug and hanging it by the handle on one of two pegs on a rack. See \autoref{fig:mug-rack} for task illustration. This task is interesting for two main reasons. First, we don’t have any CAD models for the objects. Instead, we used scanned imports of real-world objects (obtained with the ARCode app on the iPhone App Store). Second, the task has inherent multi-modality in that the mug can be hung in one of two ways for each of the two pegs.

The diffusion and residual policy system works well for this task. First, the base diffusion model captures the task's multimodality and sometimes hangs the mug on both pegs. Furthermore, the residual RL procedure keeps this multimodality intact as the base model is frozen.

\subsubsection{High-precision, Factory task: \peghole{}}
\label{app:peg-hole}

\begin{figure}[H]
    \centering
    \begin{subfigure}[b]{0.45\textwidth}
        \includegraphics[width=\textwidth]{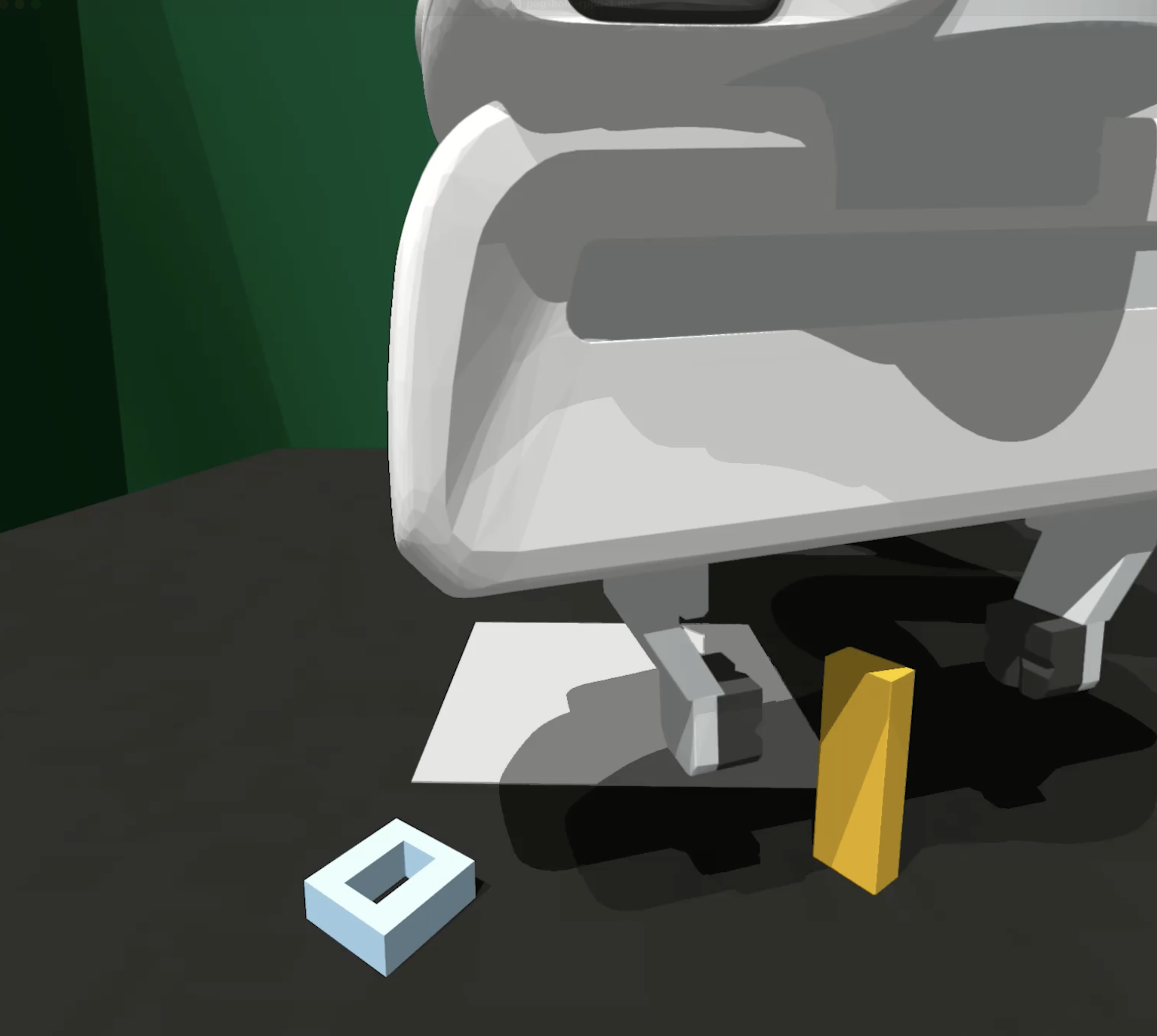}
        \caption{Example task initialization of the \peghole{} task.}
    \end{subfigure}
    \hfill
    \begin{subfigure}[b]{0.45\textwidth}
        \includegraphics[width=\textwidth]{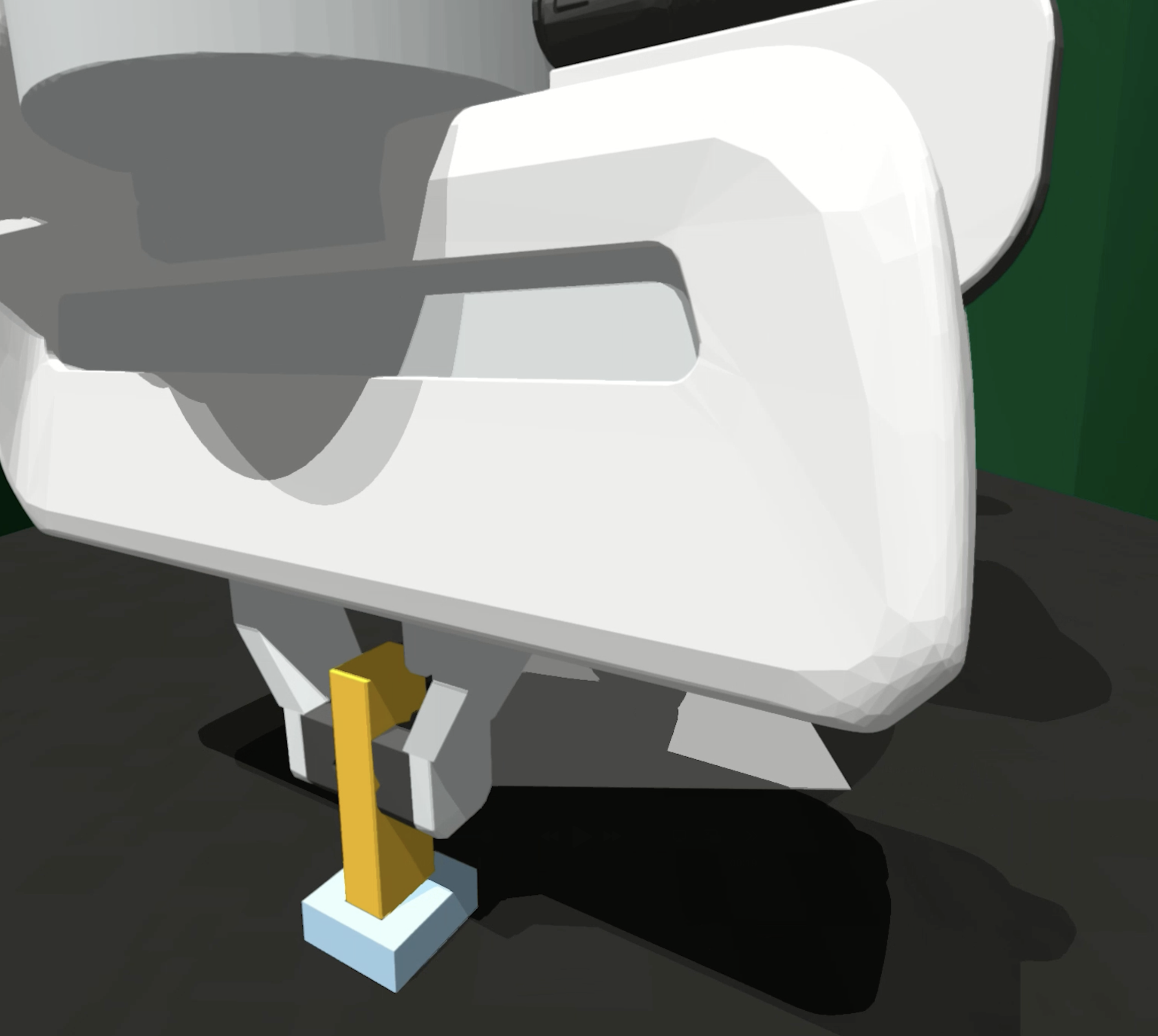}
        \caption{Example of task completion when the peg is fully inserted.}
    \end{subfigure}
    \caption{Overview of the \peghole{} task we add to push the requirement for precision. We find that the pipeline as presented works well with the same hyperparameters used for the furniture tasks.}
    \label{fig:peg-insert}
\end{figure}

To push the limits of precision in simulation, controller, and policy, we pick one of the insertion tasks from the Factory task suite~\cite{yash_factory2022_rss}, which involves grasping a peg and inserting in a hole with a 0.2mm clearance, i.e., 25x tighter than the FurnitureBench~\cite{heo_furniturebench_2023} tasks. See \autoref{fig:peg-insert} for task illustration.

Our approach also worked out of the box on this task, using the same hyperparameters as for the FurnitureBench tasks.
Here, we achieve $5\%$ success rate in pre-training and $\sim$99\% in fine-tuning. Good performance at this task is essentially entirely dominated by the ability to locally adjust the peg until it lines up with the hole, and the high final success rate achieved by our approach reflects that the local nature of the corrections learned by our residual policy is well aligned with such task scenarios.

\subsubsection{Bimanual, high-precision task: \bimaninsert{}}

\begin{figure}[H]
    \centering
    \begin{subfigure}[t]{0.45\textwidth}
        \includegraphics[width=\textwidth]{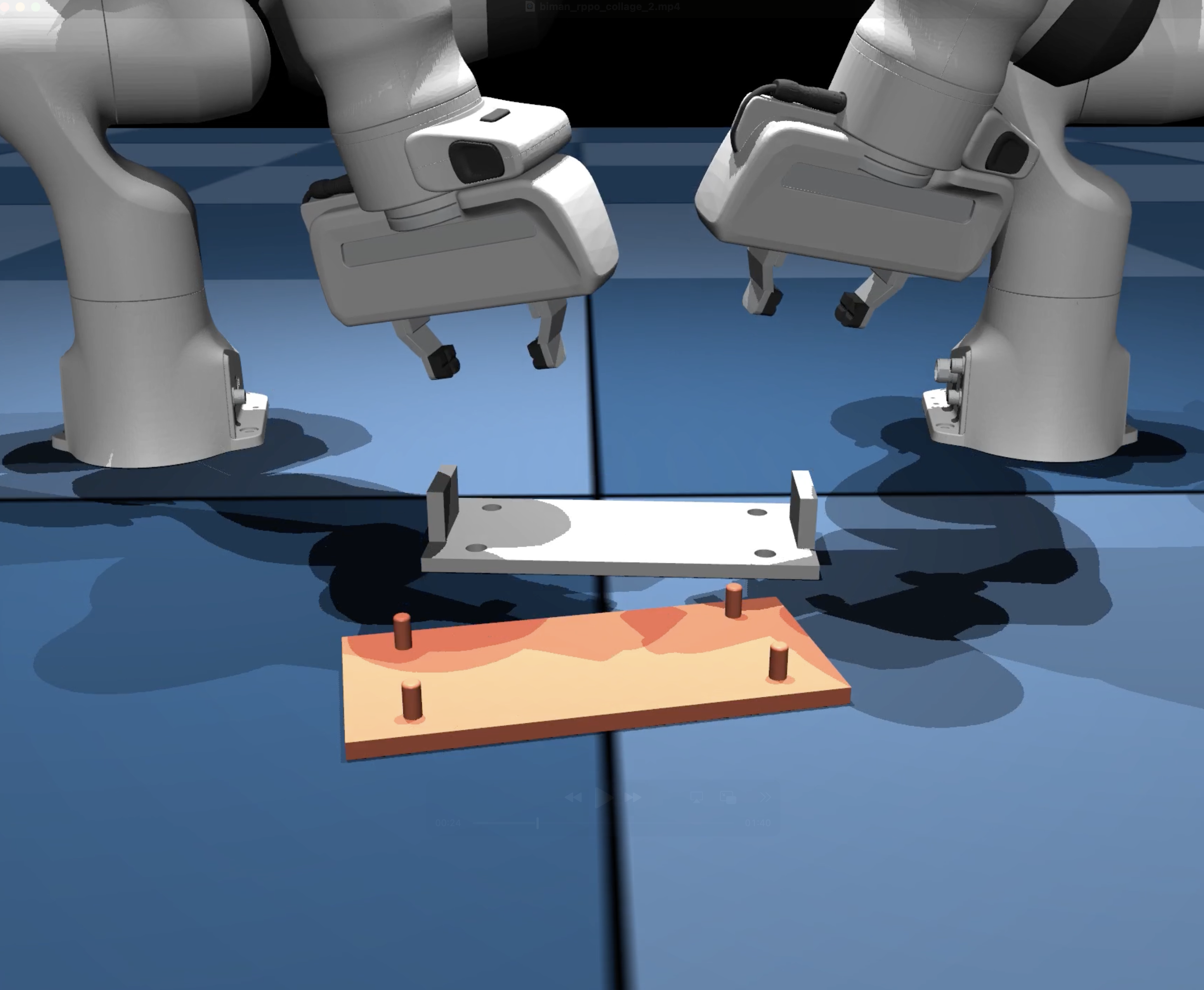}
        \caption{Example task initialization of the \bimaninsert{} task.}
    \end{subfigure}
    \hfill
    \begin{subfigure}[t]{0.45\textwidth}
        \includegraphics[width=\textwidth]{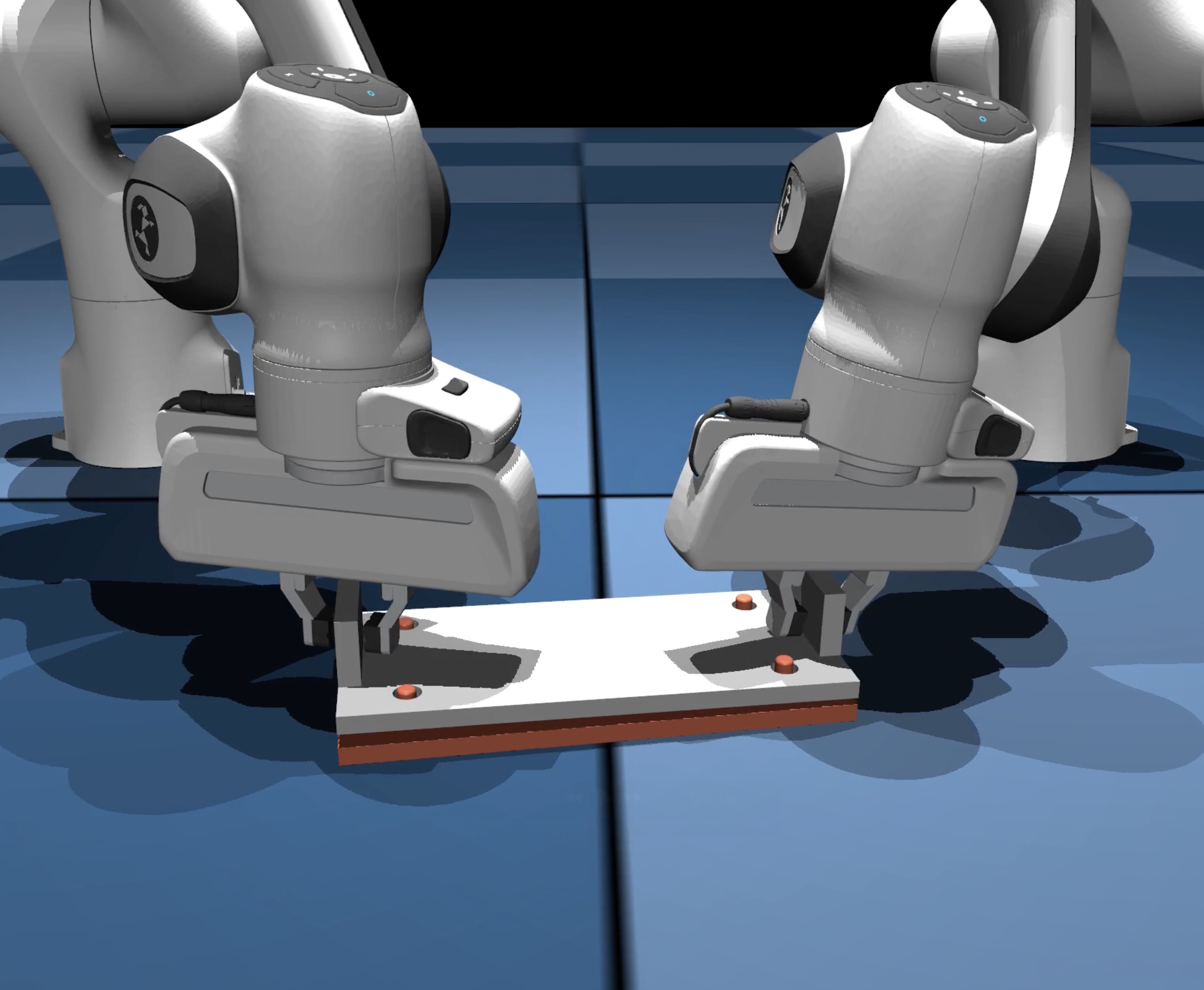}
        \caption{Example of task completion when the plate is fully inserted.}
    \end{subfigure}
    \caption{Overview of the \bimaninsert{} task we add to push the requirement for precision and bimanual coordination. We find that the pipeline as presented works well with the same hyperparameters used for the furniture tasks and that the increased action space poses no problem for mastering the task.}
    \label{fig:biman-insert-example}
\end{figure}

\begin{figure}
    \centering
    \includegraphics[width=\linewidth]{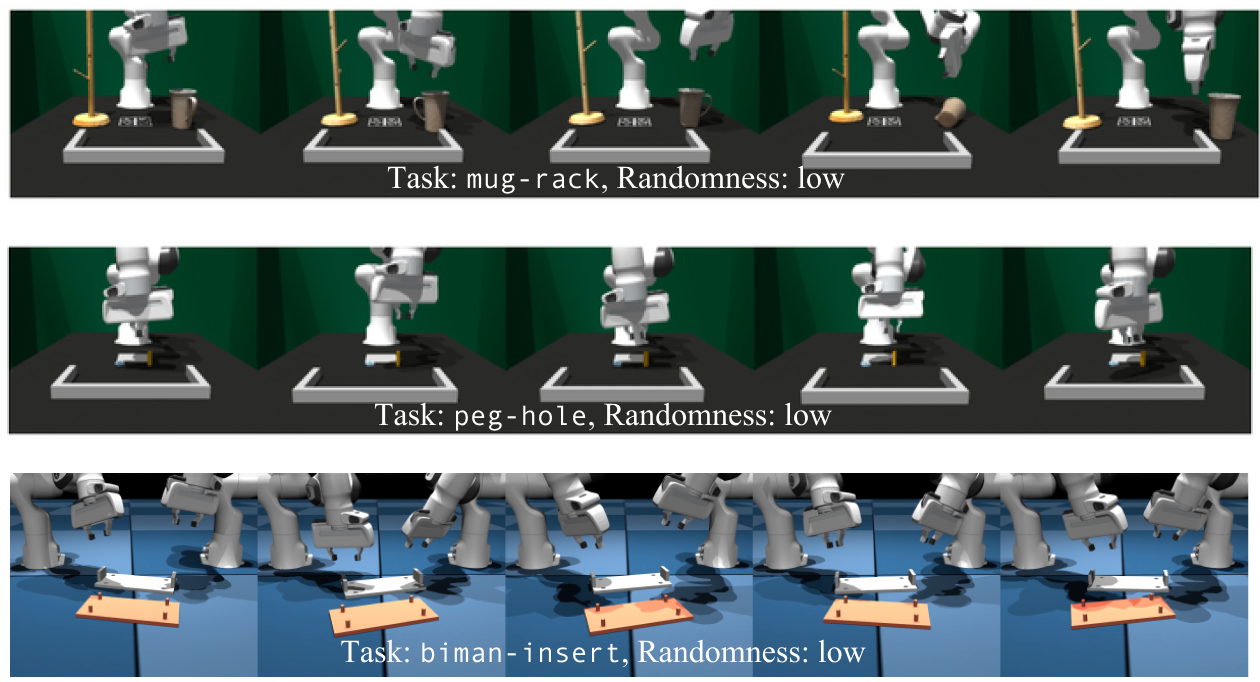}
    \caption{Examples of initial scene layouts for the 3 non-FurnitureBench tasks, \mugrack{}, \peghole{}, and \bimaninsert{}, for their default level of randomness.}
    \label{fig:randomness-levels-other}
\end{figure}

To test whether our method, \methodname{}, also works for precise tasks with larger action spaces, we create a simple bimanual industrial assembly task that we term \bimaninsert{}. See \autoref{fig:biman-insert-example} for an example initial and final state and \autoref{fig:randomness-levels-other} for several random initial states. We design the task by creating simple meshes and importing them into the MuJoCo~\cite{todorov2012mujoco} physics engine. We demonstrate the task using the augmented reality-based teleoperation interface DART~\cite{park2024dexhub}. All subsequent training uses the same code and hyperparameters as all the other tasks. This task has a relatively short horizon but has a 20-dimensional action space and relatively tight insertion tolerances. We also perform this task at 50 Hz for policy control, showing that our approach is quite general.

\begin{table}[H]
\centering
\caption{Task Attribute Overview}
\label{tab:task-attributes}
\begin{tabular}{@{}lccccccc@{}}
\toprule
& \oneleg{} & \roundtable{} & \lamp{} & \mugrack{} & \peghole{} & \bimaninsert{} \\
\midrule
Mean episode length & $\sim$500 & $\sim$700 & $\sim$600 & $\sim$150 & $\sim$200 & $\sim$400 \\
\# Parts to assemble & 2 & 3 & 3 & 2 & 2 & 2 \\
Num rewards & 1 & 2 & 2 & 1 & 1 & 1 \\
Dynamic object & \xmark & \xmark & \checkmark & \xmark & \xmark & \xmark \\
\# Precise insertions & 1 & 2 & 1 & 0 & 1 & 1 \\
\# Screwing sequences & 1 & 2 & 1 & 0 & 0 & 0 \\
Precise grasping & \xmark & \checkmark & \xmark & \xmark & \xmark & \xmark \\
Insertion occlusion & \xmark & \checkmark & \xmark & \checkmark & \xmark & \xmark \\
Control frequency & 10 Hz & 10 Hz & 10 Hz & 10 Hz & 10 Hz & 50 Hz \\
Degrees-of-Freedom & 7 & 7 & 7 & 7 & 7 & 14 \\
\bottomrule
\end{tabular}
\end{table}

\subsection{Details on randomization scheme}

The ``low'' and ``medium'' randomness settings we used for data collection and evaluation reflect how much the initial part poses may vary when the environment is reset.
We tuned these conditions to mimic the levels of randomness introduced in the original FurnitureBench suite~\cite{heo_furniturebench_2023}. However, we found that their method of directly sampling random poses often leads to initial part configurations colliding, requiring expensive continued sampling to eventually find an initial layout where all parts do not collide.

Our modified randomization scheme instead initializes parts to a single pre-specified set of feasible configurations. Then, it applies a randomly sampled force and torque to each part (where the force/torque magnitudes are tuned for each part and scaled based on the desired level of randomness). This scheme allows the physics simulation to ensure parts stay out of collision while providing a controlled amount of variation in the initial scene randomness.

\begin{figure}
    \centering
    \includegraphics[width=\linewidth]{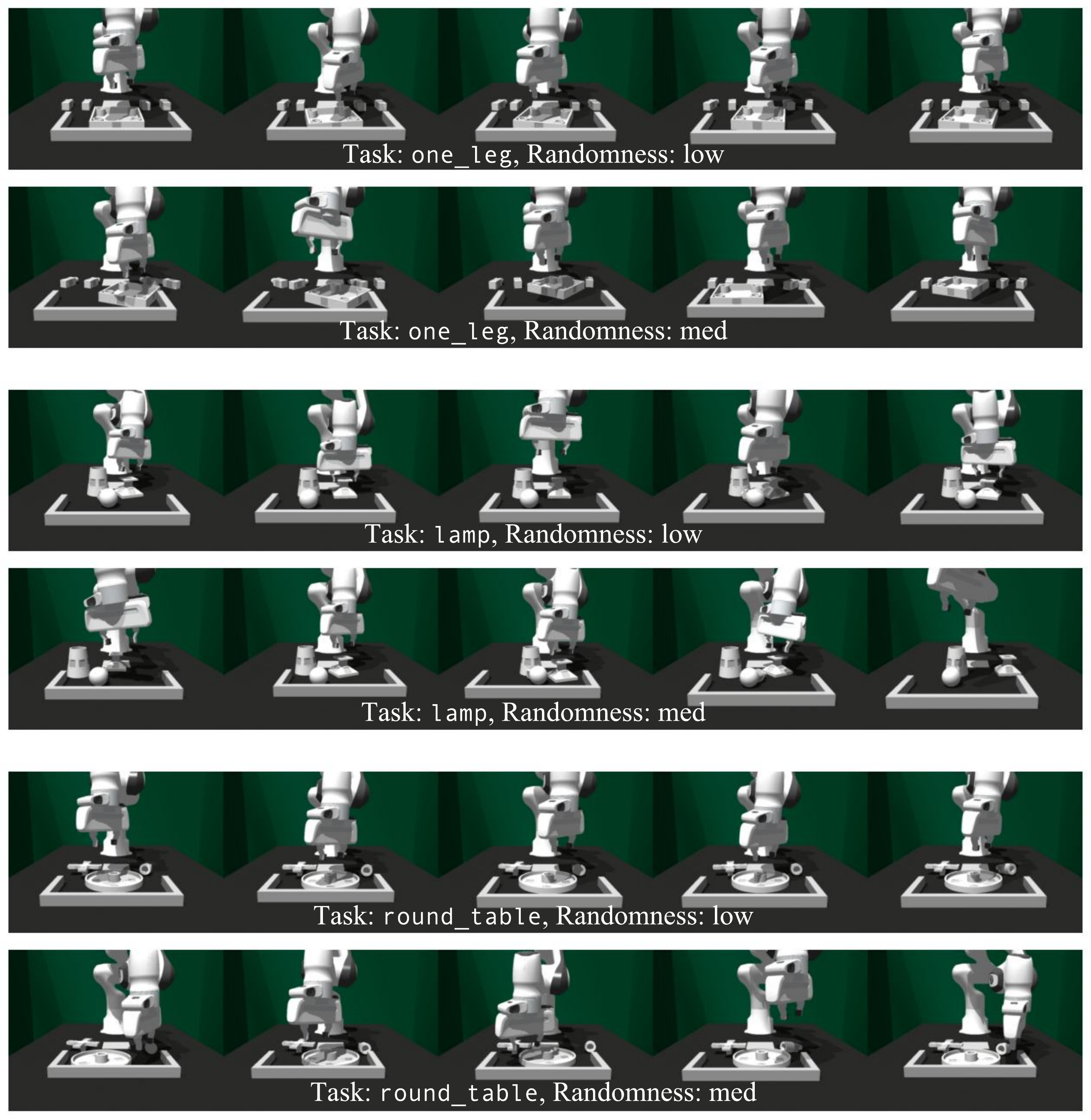}
    \caption{Examples of initial scene layouts for the tasks from the FurnitureBench task suite~\cite{heo_furniturebench_2023}, \oneleg{}, \lamp{}, and \roundtable{}, with different levels of initial part pose and obstacle fixture randomness.}
    \label{fig:randomness-levels}
\end{figure}

The second way we modified the randomization scheme was to randomize the position of the U-shaped obstacle fixture and the parts (the obstacle fixture was always kept in a fixed position in~\cite{heo_furniturebench_2023}). We reasoned that, for visual sim-to-real without known object poses, we could only imperfectly and approximately align the obstacle location in the simulated and real environment. Rather than attempting to make this alignment perfect, we instead trained policies to cover some range of possible obstacle locations, hoping that the real-world obstacle position would fall within the distribution the policies have seen in simulation. \autoref{fig:randomness-levels} shows examples of our different randomness levels for each task in simulation.

\subsection{Adjustments to FurnitureBench simulation environments}

In addition to our modified force-based method of controlling the initial randomness, we introduced multiple other modifications to the original FurnitureBench environments proposed in~\cite{heo_furniturebench_2023} to enable the environment to run fast enough to be feasible for online RL training. With these changes, we could run at a total of $\sim$4000 environment steps per second across 1024 parallel environments. The main changes are listed below:

\begin{enumerate}
    \item Vectorized reward computation, done check, robot, part, and obstacle resets, and differential inverse kinematics controller.
    \item Removed April tags from 3D models to ensure vision policies would not rely on tags to complete the tasks. We tried to align with the original levels of randomness, but only to an approximation.
    \item Deactivate camera rendering when running the environment in state-only mode.
    \item Correct an issue where the physics was not stepped a sufficient amount of time for sim time to run at 10Hz, and subsequently optimize calls to fetch simulation results, stepping of graphics, and refreshing buffers.
    \item Artificially constrained bulb from rolling on the table until robot gripper is nearby as the rolling in the simulator was exaggerated compared to the real-world parts.
\end{enumerate}

\section{RGB Sim2Real Transfer}
\label{app:sim-to-real-analysis}

\paragraph{Visualization of overlap in action space in real and sim}

\begin{figure}
    \centering
    \includegraphics[width=\textwidth]{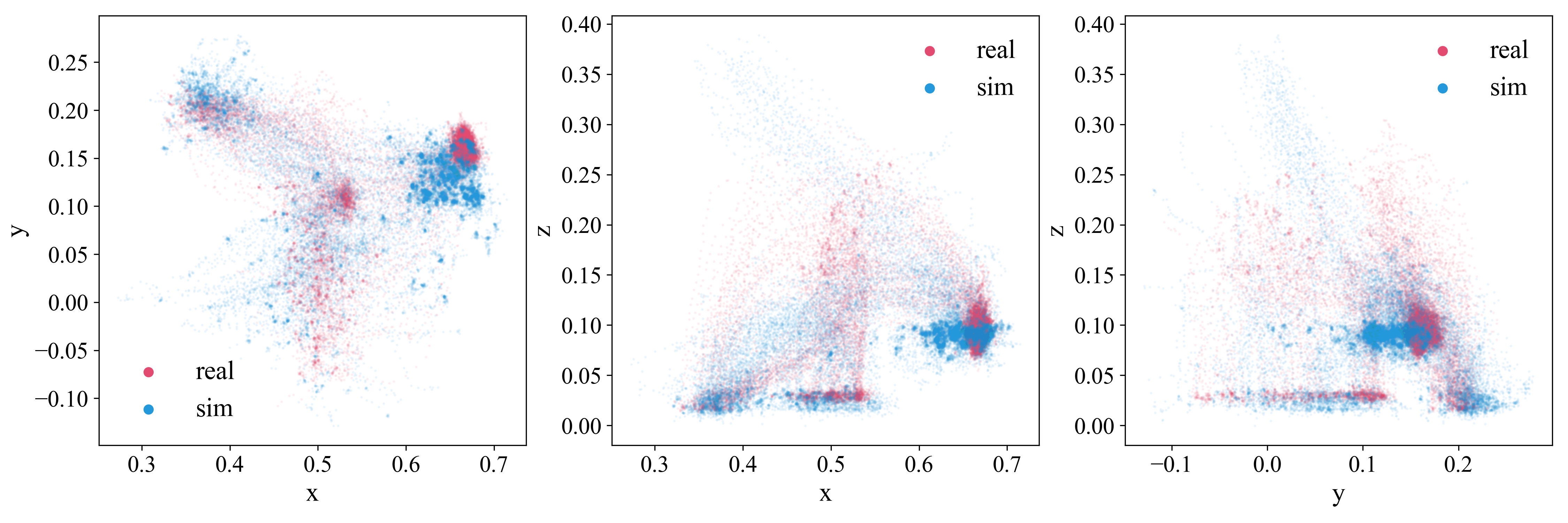}
    \caption{Plots of the $x,y,z$ action coordinates in the demo datasets for the \oneleg{} task in the real world and the simulator. That is, each dot represents one action from one of the 40/50 trajectories. Red is from real-world demos, and blue is from the simulator. \textbf{Left:} Top-down view, showing the $x,y$ positions in the workspace visited. In the top right, the insertion point is shown, where we see that the simulator has a wider distribution but could have covered better in the positive $y$-direction. \textbf{Middle:} Side-view of the actions taken in the $x,z$ plane. The insertion point is to the right in the plot; again, we see more spread in the simulation data. \textbf{Right:} Front view of the $y,z$ actions.}
    \label{fig:sim-real-action-crossection}
\end{figure}

For data from the simulation to be useful for \textit{increasing} the support of the policy for real-world deployment, we posit that it needs to \textit{cover} the real-world data. We visualize the distributions of actions in the training data in \autoref{fig:sim-real-action-crossection}. Since actions are absolute poses in the robot base frame, we can take the $x,y,z$ coordinates for all actions from simulation and real-world demonstration data and plot them. Each of the 3 plots is a different cross-section of the space, i.e., a view from top-down, side, and front. In general, we see that the simulation action distribution is more spread out and mostly covers real-world actions.

\begin{figure}
    \centering
    \includegraphics[width=\textwidth]{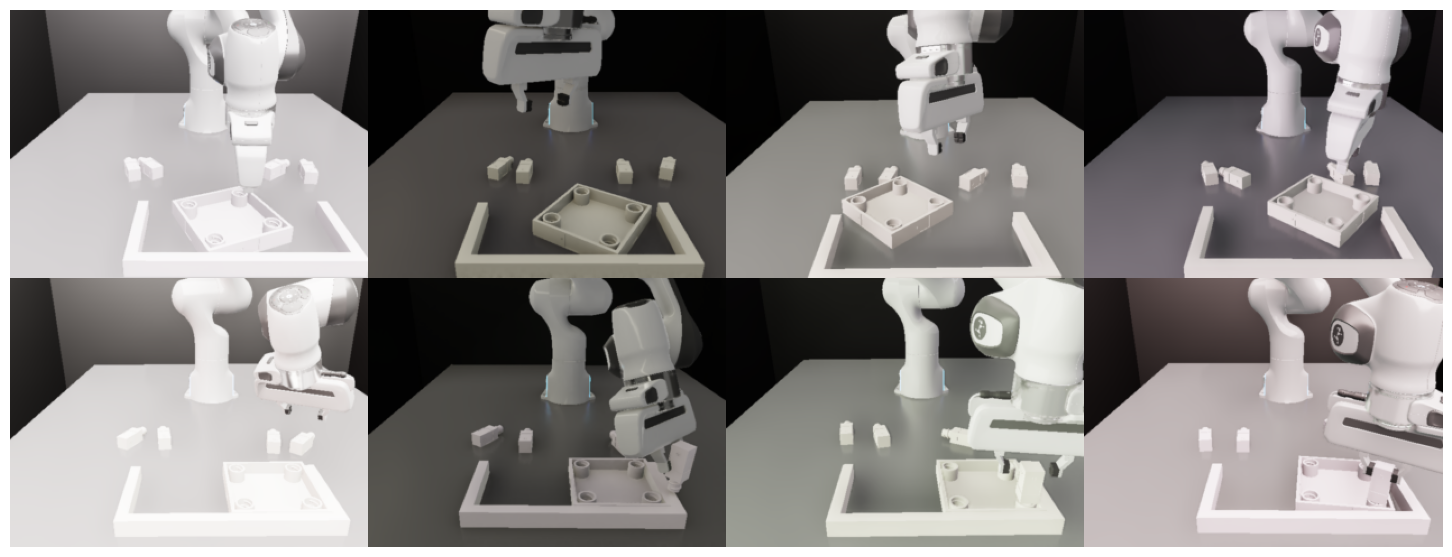}
    \caption{Examples of the randomization applied when rendering out the simulation trajectories used for co-training for the real-world policies.}
    \label{fig:render_visual_dr}
\end{figure}

\paragraph{Visual Domain randomization} In addition to randomizing part poses and the position of the obstacle, we randomize parts of the rendering which is not easily randomized by simple image augmentations, like light placement (changing shadows), camera pose, and individual part colors. See \autoref{fig:render_visual_dr} for examples of front-view images obtained from our domain randomization and re-rendering procedure.

\section{Visualization of Residual Policy Actions}

\begin{figure}
    \centering
    \includegraphics[width=0.75\linewidth]{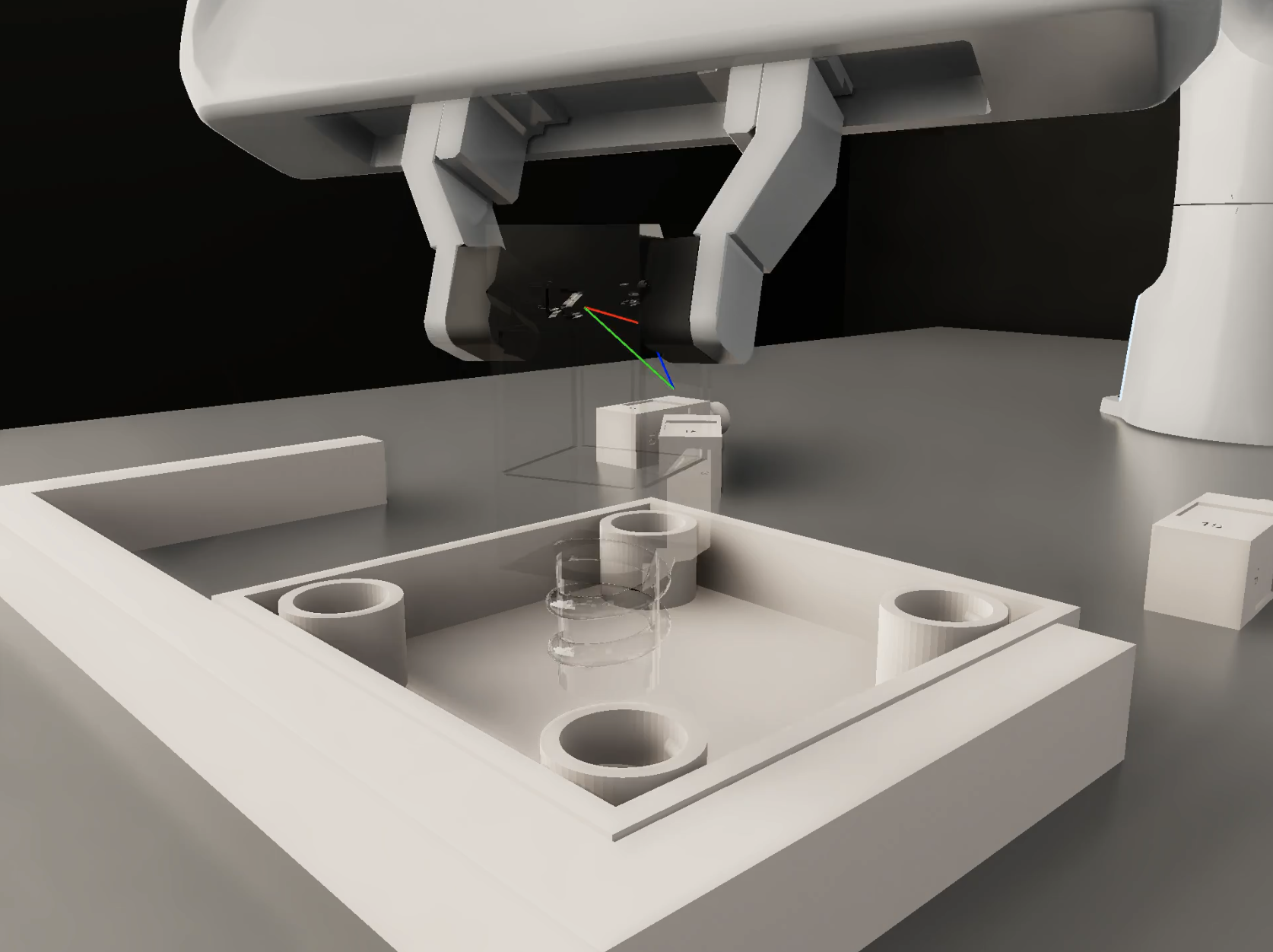}
    \caption{Visualization of the effect of the residual policy during insertion, the phase requiring the most precision. The red line shows the action commanded by the base policy. The blue is the correction predicted by the residual, and the green is the net action. The residual learns to correct actions that typically lead to failure.}
    \label{fig:residual-policy-action}
\end{figure}

We hypothesize that the strength of the residual policy is that it can operate locally and make corrections to the base action predicted by the pretrained policy operating on the macro scale in the scene. We show an example of this behavior in \autoref{fig:residual-policy-action}. Here, we visualize the base action with the red line, the correction predicted by the residual in blue, and the net action of the combined policy in green.

We find that the residual has indeed learned to correct the base policy's actions, which often leads to failure. One common example is for the base policy to be imprecise in the approach to the hole during insertion, pushing down with the peg not aligned with the hole, causing the peg to shift in the gripper, which leads to a grasp-pose unseen in the training data and the policy inevitably fails. The residual policy counteracts the premature push-down and correct the placement towards the hole, improving task success. See video examples of this behavior on the accompanying website: \url{https://residual-assembly.github.io/}.

\section{Extended Vision-Based Results and Analyses}

\subsection{Performance Impact of Distillation}
\label{app:distillation-analysis}

Next, we study how the quantity and quality of data generated by a \methodname{} policy impact the performance of vision-based student policies in real-world evaluations. We generate this data by collecting successful trajectories from the \methodname{} teacher across varied initial states and rendering corresponding camera observations. A vision-based student policy---which shares the same architecture as the teacher but includes an additional image encoder---distilled from $\sim$1,000 teacher trajectories reached 73\% success on \oneleg{}, outperforming the 50\% achieved by training the vision policy directly on human demos (see \autoref{fig:bc_vs_rl_distillation}). However, we observe a performance gap between the RL-trained \methodname{} teacher (98\%) and the distilled vision-based student (73\%), even after performance saturates with additional data. To investigate whether this gap stems from the change to visual input, we compared distillation performance between image-based and state-based students using the same number of trajectories. Their comparable performance suggests that the modality shift is not the primary cause of the performance gap. While DAgger-style online distillation might improve performance, we focused on offline distillation as it better reflects real-world deployment constraints.

Therefore, we examine the impact of the distillation dataset size. Here, we scale up the number of state-based rollouts from the trained RL policy and distill these to a state-based student. In \autoref{fig:overview} (Right), we observe that performance increases with more data, from 78\% success rate at 10k trajectories to 80\% at 100k trajectories, though not reaching the teacher policy's 98\% success rate. The same trend is evident in \autoref{fig:scaling-round-table} (though the saturation occurs earlier). These results demonstrate how simulation-based distillation can complement existing training approaches by enabling the rapid generation of large-scale synthetic datasets at a minimal cost. Beyond the data volume advantages, our RL teacher exhibits qualitatively different behaviors, such as faster movements and improved corrective actions, suggesting that this synthetic data captures valuable task strategies that could be expensive or impractical to demonstrate manually.

While DAgger~\cite{ross2011reduction} demonstrates strong sample efficiency - achieving better performance than BC with just $\sim$10k gradient steps (800 rollouts)---it requires an expert policy for online data collection. Although we could theoretically apply DAgger in simulation using our RL-trained expert policy (as demonstrated by~\cite{torne2024reconciling_rialto} for point cloud inputs), this introduces significant complexity to the pipeline and its effectiveness for RGB image-based distillation remains an open question for future work. Instead, we demonstrate that a simple approach combining offline rendering of synthetic trajectories with real-world co-training can achieve reasonable performance.

\subsection{Real-World Evaluation}
\label{app:experiments-real-world}

Finally, we evaluate the real-world performance of a sim-to-real policy trained on a mixture of a few (10/40) real-world demonstrations (\textbf{Real+Sim}) and simulation data generated by the trained residual RL policy. We compare the co-trained policy to a baseline model trained only on real-world demonstrations (\textbf{Real-Only}). We compare the success rates achieved by each policy on two sets of 10 trials for the \oneleg{} task. In the first set, we randomize part poses, while in the second set, we randomize obstacle poses (i.e., insertion location in the workspace).

We compare the co-trained policy to a baseline model trained only on real-world demonstrations. We define an evaluation grid spanning the same ranges as the low randomization setting from the FurnitureBench simulation environment. We evaluate each policy on two sets of 10 trials for the \oneleg{} task, with grid points sampling either part poses or obstacle poses (i.e., insertion location in the workspace). Each method is evaluated on the same set of grid positions to ensure fair comparison.

The results in \autoref{tab:real-world-real-cotrain} show that incorporating simulation data improves real-world performance (e.g., increasing task completion rate from 20-30\% to 50-60\%). Qualitatively, the sim-to-real policy exhibits smoother behavior and makes fewer erratic movements that might exceed the robot's physical limits. \autoref{fig:real-world-sequences} illustrates this through example trajectories: Row (A) shows successful executions where the robot completes the full assembly sequence, while Row (B) demonstrates the most common failure mode where, despite successfully grasping and transporting the parts, the policy fails to achieve precise alignment between the table leg and the hole before releasing. This misalignment failure pattern mirrors what we observe in the simulation, suggesting consistent challenges in achieving the required precision for insertion tasks.

To further probe the robustness conferred by training in simulation, we created a task variation where the part colors are changed from black to white. When rolling out the policy trained on real demos of white parts, \textbf{Real-Only}, the robot exhibited erratic behavior that caused the hardware to reach velocity limits on every trial we ran, as shown in \autoref{fig:real-black-parts} (A). When including synthetic data rendered with parts in black, the resulting policy (\textbf{Real+Sim-DR}) can perform the task again (see \autoref{fig:real-black-parts}). The resulting performance was still inferior to the performance on white parts, which motivates further work on closing the sim-to-real gap.

\subsection{Quantitative Results Failure Mode Breakdown}

\begin{figure}[H]
    \centering
    \includegraphics[width=\textwidth]{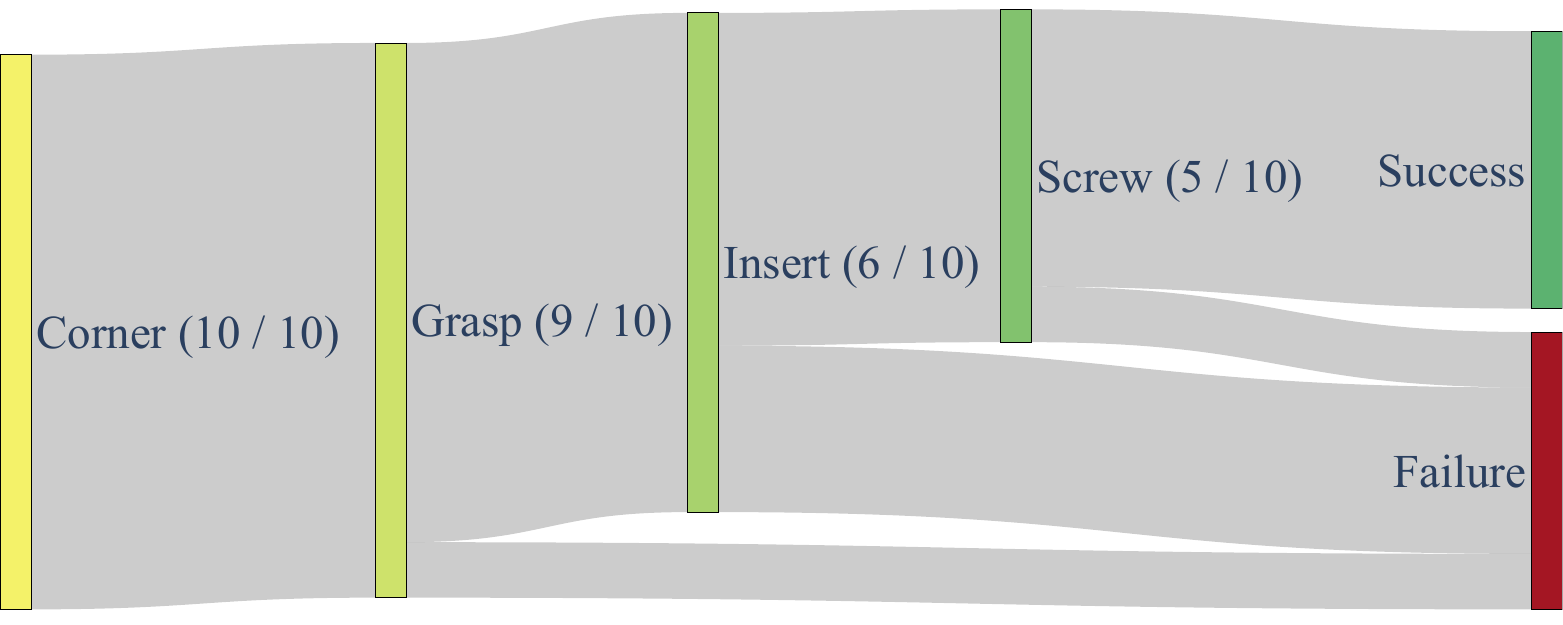}
    \caption{Sankey diagram for the success rate and failure points for the real-world rollouts with 40 real and 350 simulation demos.}
    \label{fig:real-success-sankey}
\end{figure}

The diagram in \autoref{fig:real-success-sankey} shows how successful and failed completion of individual sub-skills along the \oneleg{} task amount to our overall final success rates reported in \autoref{tab:real-world-real-cotrain} (bottom row, corresponding to ``40 real + 350 sim'' with random initial part poses and a fixed obstacle pose).

\subsection{Extension of Pipeline to Unseen Settings}
\label{app:unseen-parts}

Here, we conduct further qualitative experiments to evaluate whether our simulation-based co-training pipeline can make policies more robust to real-world parts with visual appearances that are unseen in real world demos.
To test this, we 3D printed the same set of parts used in the \oneleg{} task in black, and rolled out various policies on these black parts (rather than the white-colored parts used throughout our other experiments). 
This setting is especially relevant in industrial domains where parts can come in a variety of colors to which the assembly system must be invariant (e.g., the same piece of real-world furniture usually comes in many colors). 

\begin{figure}
    \centering
    \includegraphics[width=1.0\linewidth]{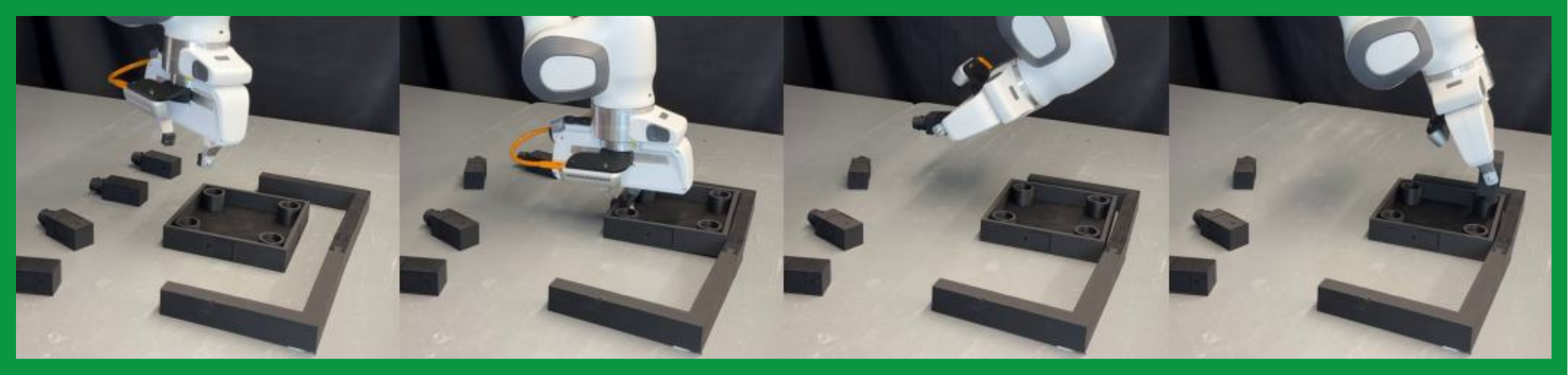}
    \caption{An example of a successful rollout of a policy co-trained on 40 real-world demos containing only white parts and 400 synthetic demos with part colors randomized.}
    \label{fig:real-world-seq-black}
\end{figure}

When deploying the policy trained on the same 40 demos as in the main experiment, which only had \emph{white}, the policy cannot come close to completing the task. The behavior is highly erratic and triggered the velocity limits of the Franka on every trial we ran.
We compare this baseline policy trained on differently colored parts to a policy co-trained on both real and synthetic data from simulation.
However, when creating the synthetic dataset for this test, we added in additional randomization of part color, with an emphasis on black or gray colors in this case, as shown in \autoref{fig:real-world-seq-black}. 
When we co-train a policy on a mix of the same real-world demos containing \emph{only} white parts as before, with a dataset of 400 synthetic demos with \emph{varying} part colors, the resulting policy can complete the task, as illustrated in \autoref{fig:real-world-seq-black} (and even when it fails at the entire task sequence, the predicted motions are much more reasonable than the erratic policy which has overfit to real-world parts of a specific color).

For example videos, please see the accompanying website: \url{https://residual-assembly.github.io/}.
We note, however, that the resulting policy is considerably less reliable than the corresponding policy rolled out with white parts, which illustrates that there is still a meaningful sim2real gap.

\begin{figure}
    \centering
    \includegraphics[width=1.0\linewidth]{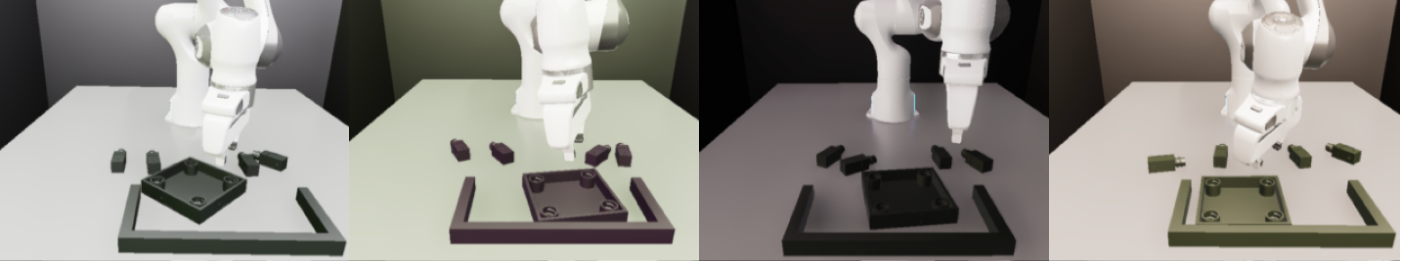}
    \caption{Randomizing the visual appearance of the scene in the simulator allows for more fine-grained control and varying attributes that are hard to isolate in standard image augmentation techniques. Here, we illustrate how we can easily cover a larger space of part appearances without jittering the colors of everything else in the scene in tandem.}
    \label{fig:dr-black}
\end{figure}

\section{Expanded Related Work}
\label{app:extended-related-work}

\paragraph{Learning robotic assembly skills}
Robotic assembly has been used by many as a problem setting for various behavior learning techniques~\cite{tang2023industreal, zhang2022learning_assembly, davchev2022residual, spector2021insertionnet1, tian2023asap}. Enabling assembly that involves multi-skill sequencing (e.g., fixturing $\rightarrow$ grasping $\rightarrow$ insertion $\rightarrow$ screwing) directly from RGB images has remained challenging, especially \emph{without} explicitly defining sub-skill-specific boundaries and supervision. Concurrent work~\cite{jiang2024transic_assembly} explores a similar framework to ours on FurnitureBench tasks~\cite{heo_furniturebench_2023}, but instead supervises learned policies on a per-skill basis and incorporates 3D point clouds.
IndustReal~\cite{tang2023industreal} also leverages RL in simulation to train high-precision skills for tight-tolerance part insertion in the real world. However, they train their RL policies from scratch using carefully-designed shaped rewards and curricula, whereas we bootstrap RL from BC pre-training, which enables RL to operate with simple sparse rewards for achieving the desired assembly. 

\paragraph{Complementary combinations of behavior cloning and reinforcement learning}
Various combinations of learning from demonstrations/behavior cloning and reinforcement learning have begun maturing into standard tools in the learning-based control development paradigm~\cite{lu2022aw_opt, Rajeswaran-RSS-18_dapg}. 
For instance, demonstrations are often used to support RL in overcoming exploration difficulty and improving sample efficiency~\cite{hu_imitation_2023_bootstrapped, torne2024reconciling_rialto, luo_serl_2024}. RL can also act as a robustification operator to improve upon base BC behaviors~\cite{lu2022aw_opt, torne2024reconciling_rialto}, paralleling the RL fine-tuning paradigm that has powered much of the recent advancement in other areas like NLP~\cite{ouyang2022training_instruct_gpt} and vision~\cite{black2023training}.
Additionally, many successful robotics deployments~\cite{chen2023visual, lee2020learning, kumar2021rma} have been powered by the ``teacher-student distillation'' paradigm, wherein perception-based ``student'' policies are trained to clone behaviors produced by a state-based ``teacher'' policy, which is typically trained via RL in simulation.
We demonstrate that our residual RL approach for fine-tuning modern diffusion policy architectures can allow each of these complementary ways to combine BC and RL to come together and enable precise manipulation directly from RGB images.

\section{Extended Limitations and Further Work}

\paragraph{Real-world distillation}

Our experiments have demonstrated the effectiveness of online learning versus offline or passive learning through behavior cloning. Still, we employ only offline learning in our teacher-student distillation phase for sim-to-real transfer, which will likely upper-bound the performance we can transfer to the real world. Combining our pipeline with techniques for online learning could improve performance significantly. However, at this point, there are significant challenges to overcome to make this practically applicable to the tasks studied herein.

The field is progressing rapidly, and we are excited to investigate how online learning in the real world can be made practical for a broader set of tasks with longer horizons and less obvious ways of performing automatic state resets in follow-up work. This effort further ties into a more general framework for pre-training and adaptation of robot systems where the deployed robot can continue learning and adapting ``on the job'' after deployment. These investigations complement the methods presented in this paper and are not in scope.

At the same time, our results indicate that making more capable systems only through increasing the collection of real-world demos may also be fundamentally limited unless online learning is introduced as a fine-tuning step in those systems.

\paragraph{Locality of online correction learning}

Though effective, we re-emphasize that our residual online reinforcement learning framework has the fundamental limitation of being bound to the pre-trained policy and mainly performing locally corrective actions. This limitation is both a strength and a weakness. First, the strong pre-trained prior allows RL to perform the tasks and improve, and having a frozen prior helps stabilize training and prevent collapse. At the same time, the degree to which online learning can generalize to states far from the training set is limited.

\paragraph{Limitations of simulators in contact-rich tasks}

We have added an experiment for a task from the Factory~\cite{yash_factory2022_rss} task suite that pushes the accuracy of the simulator more than with the original FurnitureBench~\cite{heo_furniturebench_2023} tasks. This new task has a clearance of 0.2mm for the insertion, which shows that the general BC + Residual RL framework also works well in this setting. We did not show, however, that this transfers to the real world, and it would likely be more challenging than in the original tasks for at least two reasons. First, with increased precision requirements, accurate calibration of physics parameters between the actual and simulated environment will likely matter more. Second, performing manipulation from vision when parts are smaller is more challenging.

\section{Why Action Chunking and Diffusion Policies?}

Simple feed-forward MLPs of modest size have shown impressive performance in many domains when trained with RL~\cite{kumar2021rma,lee2020learning,torne2024reconciling_rialto}, and offer a natural starting point for RL fine-tuning after BC pre-training. However, the standard MLP policies trained to directly output single action control instead of a trajectory plan through an action chunk (MLP-S) fail across all tasks we consider. Therefore, we also trained MLP policies with action chunking (MLP-C). When we introduce chunking, MLP performance improves drastically, as shown in \autoref{tab:performance_comparison}. However, we also find that the more complex Diffusion Policy (DP) architecture generally outperforms MLPs, especially in tasks of intermediate difficulty. For example, an improvement from 10\% success rate to 26\% for the \oneleg{} task on medium randomness makes subsequent fine-tuning far easier.

In one case, \lamp{} on low randomness, MLP-C outperformed DP. In qualitative evaluations, we find that DP has smoother and faster actions, which is generally beneficial. Still, it seems to hurt performance in this case, as it tends to retract before the gripper fully grasps the lamp base. We also find that all methods struggle with the most challenging tasks, on which MLP-C and DP both achieve less than 5\% success rate, indicating that there is still room for improvement in BC methods. The \peghole{} task, despite its relatively short horizon of $\sim$100 timesteps, proved particularly challenging for BC methods. This task involves a $\sim$0.2 mm tolerance insertion, resulting in a 5\% success rate. This poor performance on a short yet precise task lends credence to the hypothesis that BC methods are ill-equipped to handle high-precision requirements.


\section{Further Analysis of Offline versus Online Learning}

\subsection{Distillation scaling analysis}

\begin{figure}[H]
    \centering
    \begin{subfigure}[b]{0.45\textwidth}
        \includegraphics[width=\textwidth]{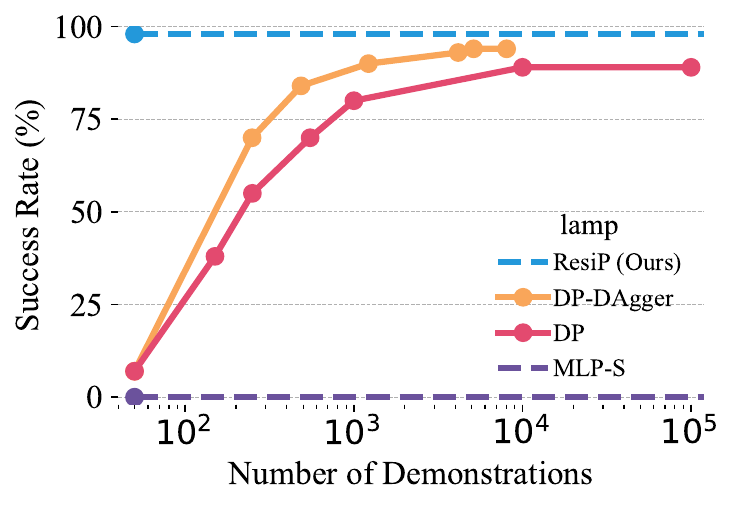}
        \caption{Scaling analysis for the \lamp{} task. This task appears significantly more conducive to offline learning than the other tasks tested.}
    \end{subfigure}
    \hfill
    \begin{subfigure}[b]{0.45\textwidth}
        \includegraphics[width=\textwidth]{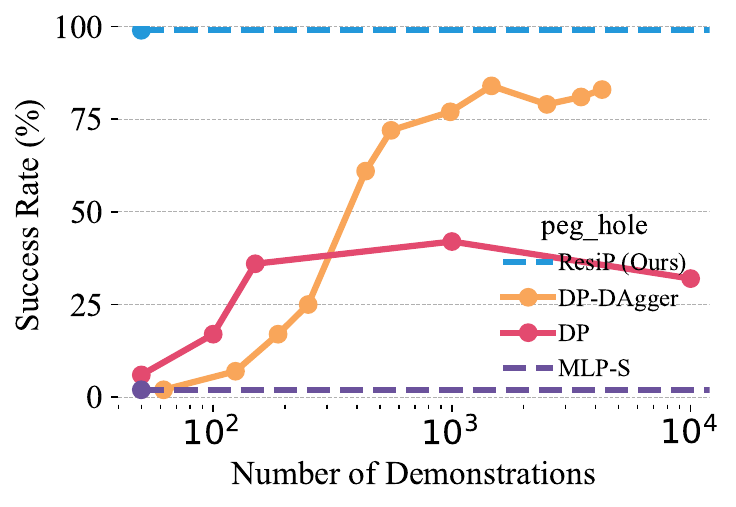}
        \caption{Success rates in exploration phase of training for \texttt{one\_leg}, low randomness.}
    \end{subfigure}
    \caption{We run similar scaling analyses as in \autoref{fig:overview} (right) and \autoref{fig:scaling-round-table} for \lamp{} and \peghole{}. The general findings of interactive learning are that it is more efficient and has higher asymptotical performance. However, the difference appears to be much smaller for \lamp{} and bigger for \peghole{}. What drives these differences are left for future work.}
    \label{fig:distillation-scaling-app}
\end{figure}

The scaling analyses in \autoref{fig:distillation-scaling-app} show the same trends as in \autoref{fig:overview} (right) and \autoref{fig:scaling-round-table} for the tasks \lamp{} and \peghole{}. We believe that this data point suggests that pure offline learning from demonstrations may not be sufficient for policies to learn robust and reactive policies, pointing towards the necessity for techniques like RL to reach a high level of robustness and reliability.

\subsection{Interactive distillation with DAgger}

DAgger~\cite{ross2011reduction} can learn from scratch significantly more efficiently than pure BC measured in both gradient steps and samples. We have added the DAgger performance to the scaling plot, shown in orange in \autoref{fig:distillation-scaling-app}.
Consider the scaling plot in \autoref{fig:overview} (right) as an example.
In $\sim$10k gradient steps, DAgger surpasses BC from 50 human demos trained with $\sim$100k steps. After 10k steps, it has around 800 rollouts in the aggregated dataset. After around 20k gradient steps, it seems to surpass the best-performing BC distillation runs using more than 10k rollouts and 500k gradient steps, at which point it has $\sim$1.5k demonstrations in the replay buffer.
This result highlights the effectiveness of online and interactive learning as opposed to learning purely passively from an offline dataset. Furthermore, it highlights that the expert we query is an effective teacher. It also highlights that for interactive learning to be effective, one needs to have a teacher ready to be queried as learning progresses.

\section{Residual RL ablations}

\subsection{Effect of fully versus partially closed-loop policies}
\label{app:closed-loop-ablation}

One differentiating factor of our residual model from some prior work is that the base and residual models make predictions at different frequencies, i.e., every 8 timesteps for the base model and every timestep for the residual model. Making predictions with the most up-to-date information is likely an easier prediction problem, and we expect this to work better than the ``standard'' setup of letting the residual correct the full output of the base model. When training a residual model that corrects a whole chunk at a time but otherwise uses the same hyperparameters, we observe that training is less sample efficient and performance saturates at a lower success rate. In particular, the chunked residual policy reaches $\sim$85\% success rate in about 250 million environment steps, while the one-step residual needs about 75 million.

\begin{figure}
    \centering
    \includegraphics[width=0.5\linewidth]{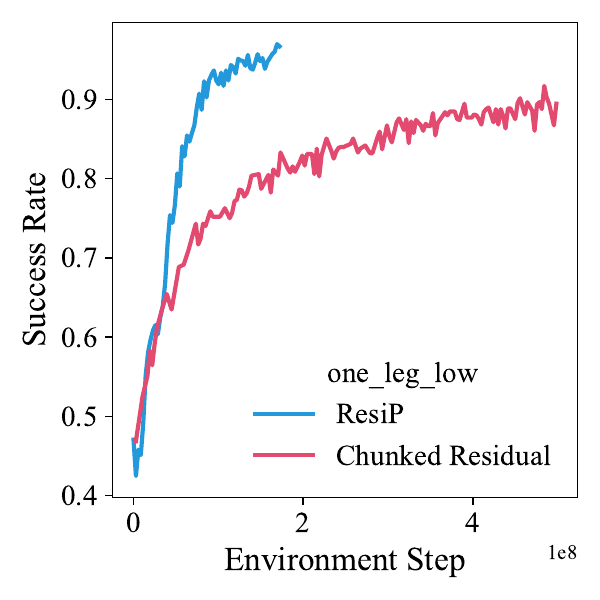}
    \caption{When learning a residual correction term that corrects the whole chunk at a time online, we find that the learning is significantly slower (measured in environment steps) and saturates a lower asymptotic level.}
    \label{fig:chunked-residual-ablation}
\end{figure}

To further probe the difference between fully closed-loop policies and those using chunking, we evaluate the policies with perturbations added to the parts in the environment throughout the episode. In particular, at each timestep, 1\% of parts across the environments will have a random force applied to them. The forces are sampled from the same distribution as the initial part randomization distribution.

See \autoref{fig:online_perturbation_comparison} and \autoref{tab:success_rates_perturbations} for results. We generally see that the partially open-loop policies have a bigger drop in performance when perturbations are introduced, around 20 percentage points compared to 12 for the one-step residual model.

\begin{table}[H]
  \centering
  \begin{tabular}{lccc}
    \toprule
    Model & No Perturb & W/ Perturb & Drop in SR \\
    \midrule
    Standard RPPO & 98\% & 86\% & 12 pp \\
    Chunked RPPO & 92\%  & 73\% & 19 pp \\
    Chunked pre-trained BC & 52\% & 32\% & 20 pp \\
    DAgger chunked student DP & 90\% & 68\% & 24 pp \\
    \bottomrule
  \end{tabular}
  \vspace{12pt}
  \caption{Success rates with/without perturbations for different models. SR = Success Rate, pp = percentage points.}
  \label{tab:success_rates_perturbations}
\end{table}

\subsection{Residual base policy ablation}
\label{app:residual-base-ablation}

\begin{figure}[H]
    \centering
    \begin{subfigure}[b]{0.30\textwidth}
        \includegraphics[width=\textwidth]{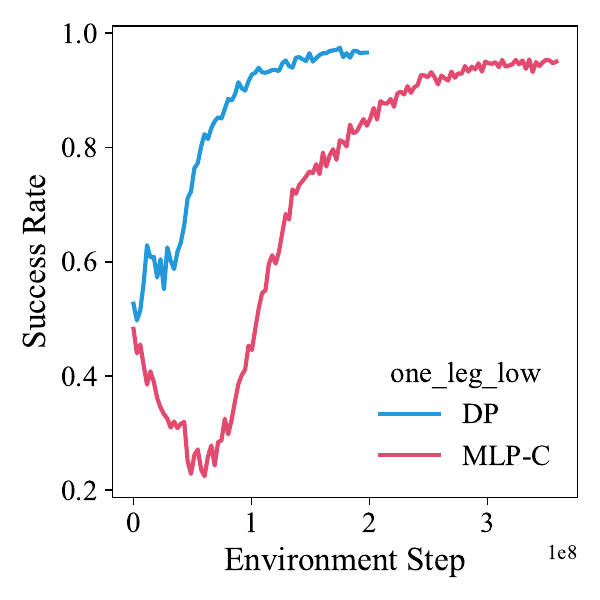}
        \caption{Evaluation success rates for RL training for \texttt{one\_leg}, low randomness for Diffusion and MLP base policy.}
        \label{fig:residual-base-mlp-low}
    \end{subfigure}
    \hfill
    \begin{subfigure}[b]{0.30\textwidth}
        \includegraphics[width=\textwidth]{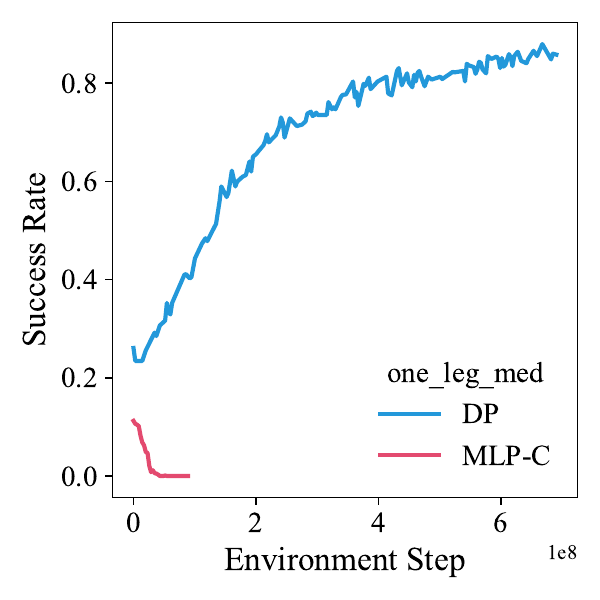}
        \caption{Evaluation success rates for RL training for \texttt{one\_leg}, medium randomness for Diffusion and MLP base policy.}
        \label{fig:residual-base-mlp-med}
    \end{subfigure}
    \hfill
    \begin{subfigure}[b]{0.30\textwidth}
        \includegraphics[width=\textwidth]{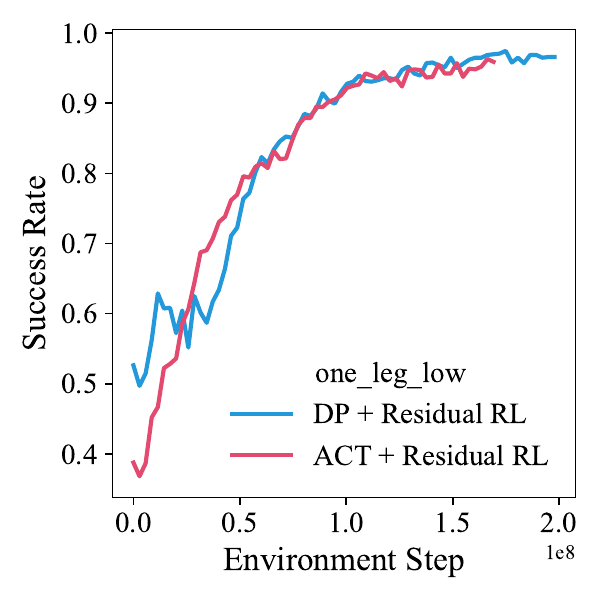}
        \caption{Evaluation success rates for RL training for \texttt{one\_leg}, low randomness for Diffusion and ACT base policy.}
        \label{fig:residual-base-act}
    \end{subfigure}
    \caption{We compare the diffusion-based residual RL training performance with the best-performing MLP as the base model in (a) and (b). As we can see, despite having similar pertaining performance (for low randomness), the MLP-based residual model performs poorly compared to the diffusion-based one. On a higher randomness setting, it fails to complete the task. We compare with using the ACT~\cite{zhao_rss23_aloha} as the base policy in (c). The pre-training performance is slightly worse, but performance quickly catches up during online learning, indicating that the ACT model is also well-suited for residual learning.}
    \label{fig:residual-base-ablation}
\end{figure}

\paragraph{MLP as base}
To further tease apart what part of the diffusion policy that provides the most important performance increase, the action chunking or the denoising diffusion process, we run the same residual PPO run for the \texttt{one\_leg} task as before, but with the best-performing BC MLP model in place of the diffusion policy. The results are shown in \autoref{fig:residual-base-mlp-low} and \autoref{fig:residual-base-mlp-med}.
The resulting training dynamics are intriguing. Despite the initial success rate of the base model being close to that of the diffusion model, the success rate drops markedly when exploration noise is introduced. This is especially visible in the training performance in plot 2 below. We also notice that the evaluation performance drops as the residual model explores and learns more. However, the residual is eventually able to find actions that the MLP responds better to and, in the end, converges to a similar performance as the diffusion-based runs.
In the more challenging task with higher initial state randomness, the same initial dynamic plays out, but the training performance drops to zero, causing the learning to collapse.
We conclude that any base model achieving a high enough initial success rate can be plugged into our framework (and, based on our BC experiments, a base model with chunking is likely to outperform one without chunking) but that the expressivity and robustness to input noise offered by diffusion de-noising also contributes to downstream performance benefits during residual RL.

\paragraph{ACT as base}

To see if the robustness to noise and suitedness for residual learning is unique to the diffusion-type model, we also implement and test using Action-Chunked Transformer (ACT)~\cite{zhao_rss23_aloha} as the base model for the \oneleg{} task at low randomness.
With some tuning, we find that the ACT model can achieve comparable performance as the diffusion model, though slightly lower, in pre-training. In the fine-tuning phase, however, it functions as well and stably as the diffusion model base, as shown in \autoref{fig:residual-base-act}.
This suggests that the residual RL framework is suitable for a wide range of fine-tuning applications and may be applied to fine-tune even larger and possibly multi-task models.

\subsection{Residual action scaling parameter ablation}
\label{app:alpha-ablation}

A design choice we make is the parameter $\alpha=0.1$. The parameter choice is somewhat arbitrary and was informed by some intuitions about the task. For example, since the residual model intends to make local corrections, we want to imbue it with that inductive bias. In the normalized action space, the workspace is constrained to [-1, 1], and letting a $\sigma=1$ for the residual Gaussian model correspond to [-0.1, 0.1] on the macro scale seemed reasonable.

We have tested more values of the parameter $\alpha\in\{0.01, 0.05, 0.2, 1.0\}$, but kept the resulting exploration noise on the macro scale fixed (i.e., scaled with the value of $\alpha$, so $\alpha_1 \sigma_1 = \alpha_2 \sigma_2$). The result, shown in the figure below, shows a remarkable robustness to this parameter, and all cases have very similar performance.

We note a couple of observations. First, $\alpha=0.2$ seems to perform slightly better than our original $\alpha=0.1$. Second, different levels of $\alpha$ also result in very different magnitudes of activations at the last layer, which impacts losses. This experiment shows that the resulting performance does not differ significantly, but we suspect it could make training less stable in harder settings.

\begin{figure}[H]
    \centering
    \includegraphics[width=0.98\linewidth]{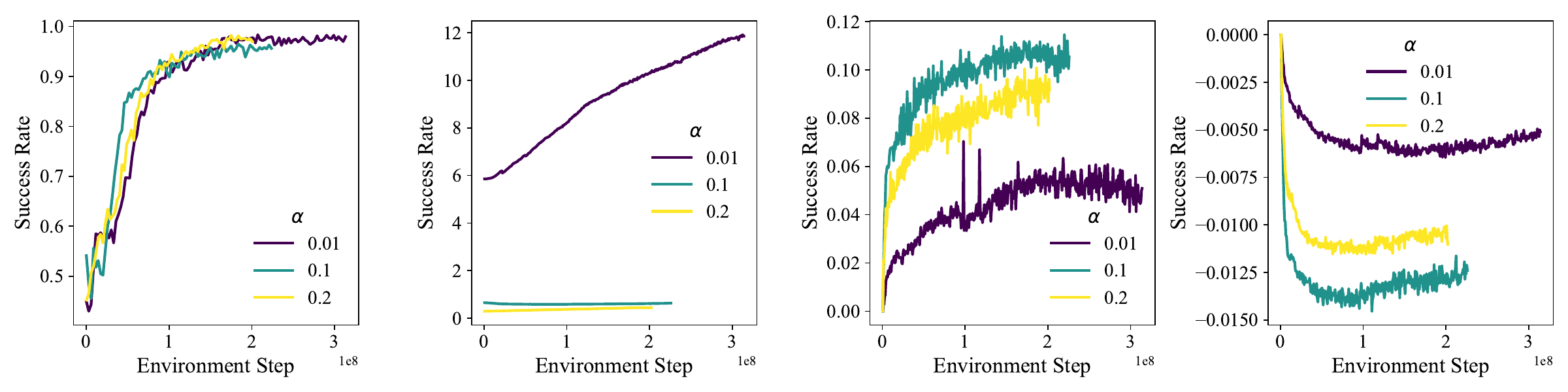}
    \caption{We test different values of the residual action scaling parameter $\alpha$ and test it for values $\alpha\in\{0.01, 0.1, 0.2\}$ while adjusting the exploration noise to be such that the macro-level exploration is the same initially. We find that for success rates in this task, the value is not crucial but does cause training dynamics to change, particularly the residual model output norms and policy loss magnitude.}
    \label{fig:alpha-ablation}
\end{figure}

\end{appendices}

\end{document}